\documentclass[runningheads]{llncs}

\usepackage{eccv}

\usepackage{eccvabbrv}

\usepackage{graphicx}
\usepackage{booktabs}
\usepackage{multirow}
\usepackage[table]{xcolor}
\definecolor{cvprblue}{rgb}{0.21,0.49,0.74} % 在表格前或导言区定义颜色
\usepackage[ruled,vlined]{algorithm2e}

\usepackage[accsupp]{axessibility}  % Improves PDF readability for those with disabilities.

\usepackage{hyperref}

\usepackage{orcidlink}

\begin{document}

\title{CMDS-AD: Cross-Modal Dual-Stream Decoupling for Few-Shot Anomaly Detection}

\titlerunning{CMDS-AD for Few-Shot Anomaly Detection}

% Second Author\inst{2,3}\orcidlink{1111-2222-3333-4444} \and
% Third Author\inst{3}\orcidlink{2222--3333-4444-5555}}

\author{Junhao Cai\inst{1} \and
Junyu Chen\inst{2} \and
Deyu Zeng\inst{1, 2}\thanks{Corresponding author.} \and
Junhao Pang\inst{1} \and
Qiwei Liang\inst{1, 3} \and
Xiaopin Zhong\inst{1} \and
Zongze Wu\inst{1}}

\authorrunning{J.~Cai et al.}

% Springer Heidelberg, Tiergartenstr.~17, 69121 Heidelberg, Germany

\institute{Shenzhen University, Shenzhen, Guangdong 518060, China \\
% 深大相关的邮箱（包含 Qiwei Liang 的深大后缀邮箱）
\email{caijunhao27@gmail.com, 2500092013@mails.szu.edu.cn, liangqiwei2022@email.szu.edu.cn, \{xzhong,zzwu\}@szu.edu.cn} \and
Guangzhou Maritime University, Guangzhou, Guangdong 510725, China \\
% 广航相关的邮箱
\email{\{zengdeyu,chenjunyu\}@gzmtu.edu.cn} \and
Hong Kong University of Science and Technology (Guangzhou), Guangzhou, Guangdong 511453, China}

\maketitle

\begin{abstract}
Few-shot anomaly detection remains challenging due to limited training data. Multi-modal anomaly detection (MAD) offers a viable solution, leveraging 3D geometric cues to enrich 2D RGB representations and compensate for this scarcity. However, existing MAD methods apply spatially uniform feature processing, conflating stable macroscopic structures with high-frequency localized defect signals, exacerbating cross-modal misalignment and inflating false-positive rates. To overcome this, we present CMDS-AD, a Cross-Modal Dual-Stream Anomaly Detection framework. A LoRA-guided diffusion model generates diverse RGB samples to mitigate extreme data scarcity. For 3D normal augmentation, we employ a pre-trained diffusion model as a normal estimator. Crucially, this estimator inherently acts as a non-linear low-pass filter, directly extracting low-frequency normal representations from RGB inputs. This establishes an auxiliary estimated stream of purely low-frequency information, anchoring robust structural templates and assisting the uncompressed real stream, containing coupled high- and low-frequency components, to precisely isolate micro-defects. A Coordinate-Aware Hierarchical Feature Mapper adaptively aligns cross-modal semantics, while a multiplicative scoring mechanism filters modality-specific noise. Under the extreme 1-shot setting, CMDS-AD achieves absolute performance gains of \textbf{5.7\%} (I-AUROC) and \textbf{2.0\%} (AUPRO) on MVTec 3D-AD, alongside \textbf{7.7\%} and \textbf{5.6\%} improvements on EyeCandies, establishing a new state-of-the-art. Code is available at \href{https://github.com/Junhaocai27/CMDS-AD}{\texttt{Junhaocai27/CMDS-AD}}.
\end{abstract}

\section{Introduction}
\label{sec:intro}

Anomaly detection (AD) has become an indispensable component of industrial visual inspection, ensuring product quality and manufacturing stability where manual inspection is inefficient and error-prone. Since anomalous samples (\eg defects, scratches, structural damages) are unpredictable and extremely rare in real-world scenarios, most existing AD methods adopt an unsupervised paradigm, training exclusively on normal, defect-free data to identify anomalies as deviations from nominal distributions \cite{deng2022anomaly, ristea2022self, tien2023revisiting, zhang2023destseg}.

\begin{figure}[t!]
    \centering
    % 使用 resizebox 包裹整个子图区域。
    % 如果你想整体缩小图片，可以把 \textwidth 改成 0.9\textwidth 等数值
    \resizebox{0.91\textwidth}{!}{%
        % 使用 minipage 包装，确保 \hfill 在 resizebox 内部依然能正常撑开间距
        \begin{minipage}{\textwidth}
            \centering
            % Subfigure (a) - Mechanism Diagram
            \begin{subfigure}[b]{0.5\columnwidth} % Scaled down relative width
                \centering
                \includegraphics[width=\textwidth]{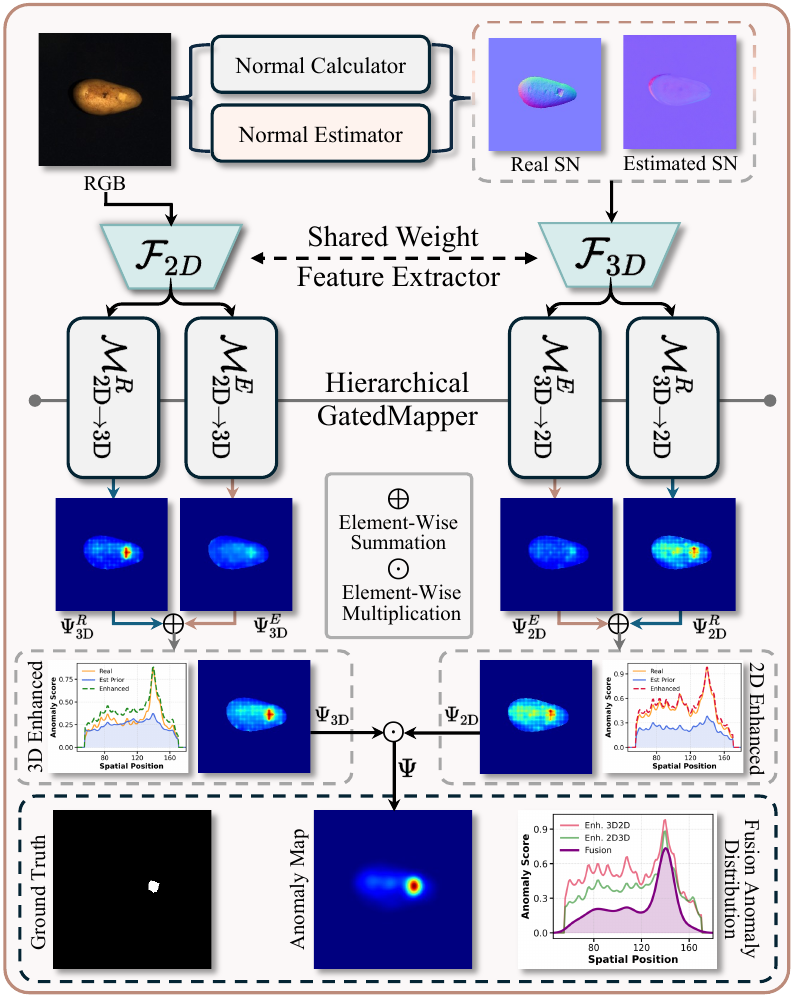}
                \caption{Mechanism}
                \label{fig:intro_a}
            \end{subfigure}
            \hfill % Adds horizontal fill between containers
            % Subfigure (b) - Original Radar Plot (scaled down)
            \begin{subfigure}[b]{0.43\columnwidth} % Scaled down relative width
                \centering
                \includegraphics[width=\textwidth]{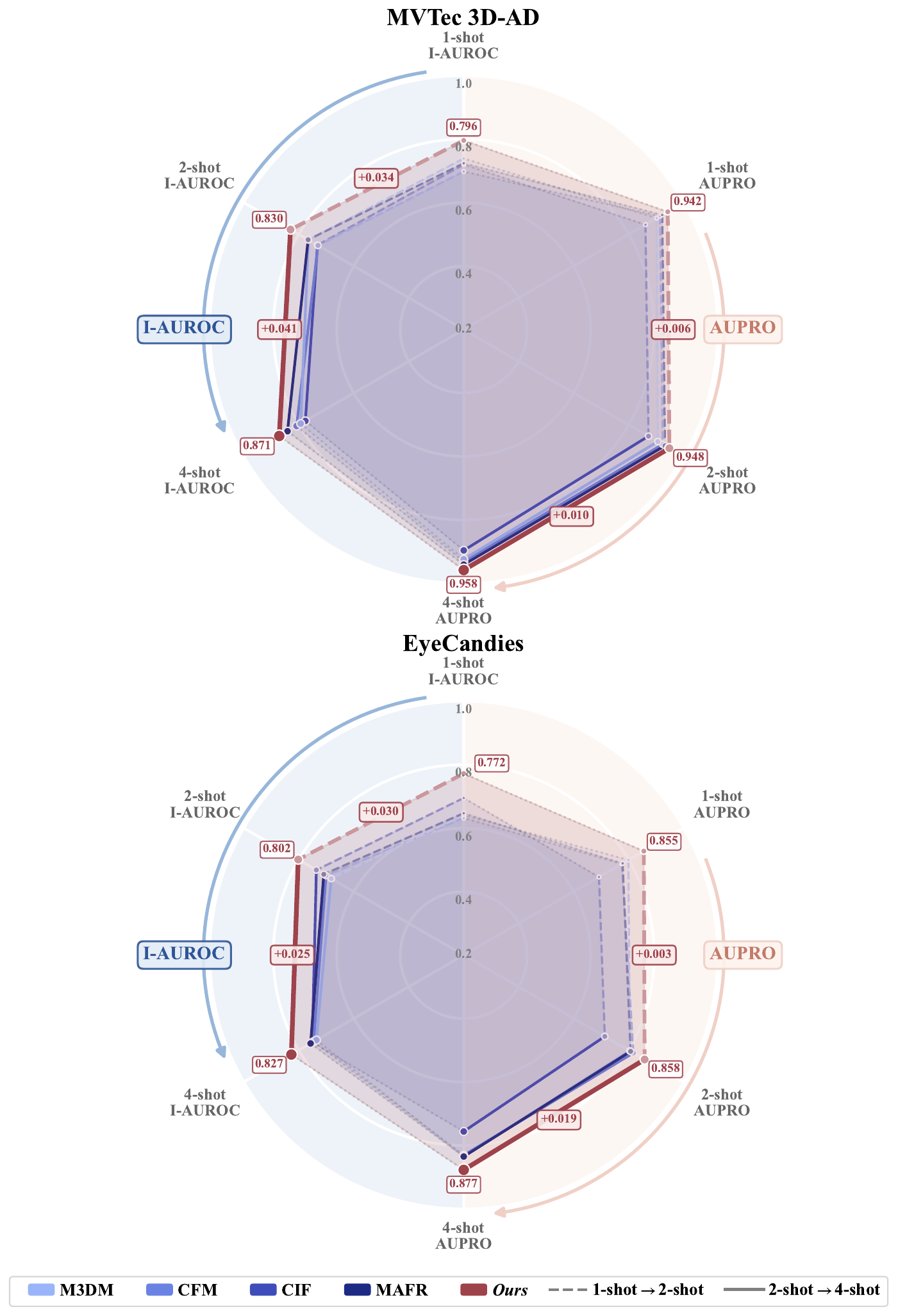}
                \caption{Results}
                \label{fig:intro_b}
            \end{subfigure}
        \end{minipage}%
    }
    % Main combined caption (放在 resizebox 外面，防止标题文字跟着一起变形或缩小)
    \caption{Few-shot 3D anomaly detection overview. (a) Illustration showing how estimated streams enhance defect representation in real streams, and multi-modal 2D-3D multiplication further localizes defects. (b) Radar plots on MVTec 3D-AD and EyeCandies report few-shot (1, 2, 4-shot) I-AUROC and AUPRO. Our method (\textcolor{red!70!black}{red}) outperforms baselines (\textcolor{blue!70!black}{blue}) with annotated gains, demonstrating robust improvements.}
    \label{fig:intro}
\end{figure}

With the rapid advancement of 3D sensors, multi-modal anomaly detection (MAD) has garnered increasing interest \cite{li2024towards, long2025revisiting, yan2022let}. Relying on a single modality (\eg RGB) often fails to capture crucial geometric cues obscured by complex surface textures, whereas 3D information provides highly complementary structural insights \cite{liang2025look, zavrtanik2024cheating}. While significant progress has been made in MAD under full-shot settings—ranging from memory-based pattern matching \cite{chu2023shape, wang2023multimodal} and feature adaptation \cite{sui2025incomplete, tu2024self} to diffusion-driven reconstruction \cite{liu2024learning, zhou2024r3d}—existing approaches rely on a large volume of normal training data. In agile manufacturing, acquiring extensive multi-modal normal samples for every new product line is highly expensive. Consequently, few-shot anomaly detection (FSAD) has emerged as a critical alternative \cite{fang2023fastrecon, huang2022registration, jeong2023winclip, lee2024anople, tian2024foct}. However, when restricted to severe few-shot scenarios (\eg 1 to 4 samples), the performance of state-of-the-art multi-modal methods drops precipitously (as evidenced in Fig.~\ref{fig:intro}).

% Requires \usepackage{graphicx}, \usepackage{xcolor}, and \usepackage{subcaption} in your preamble

The fundamental bottleneck of extending MAD to few-shot scenarios is the profound modality gap and modality interference under data scarcity. Existing methods attempt to bridge this gap through various fusion strategies, but often fall short. For instance, fusing modalities via input-level concatenation (\eg AST \cite{rudolph2023asymmetric}) or multi-level spatial fusion (\eg MMRD \cite{gu2024rethinking}) risks severe modality interference. Cross-modal mapping approaches, such as CFM \cite{costanzino2024multimodal}, attempt to map features between modalities; however, the inherent modality gap often leads to large reconstruction errors even for normal samples, inherently resulting in high false-positive rates and coarse localization boundaries. Meanwhile, memory-based architectures like M3DM \cite{wang2023multimodal} and ShapeGuided \cite{chu2023shape} struggle because limited normal samples fail to adequately cover the nominal feature patterns. Fundamentally, these methods treat all spatial features uniformly, neglecting the frequency properties of industrial data. By entangling stable low-frequency macroscopic structures with unpredictable high-frequency localized variations (\eg minor texture shifts), they inevitably confuse normal sensor noise with genuine structural defects in data-scarce scenarios.

To overcome these severe limitations, we propose Cross-Modal Dual-Stream Anomaly Detection (CMDS-AD), a novel few-shot MAD framework that explicitly establishes a dual-stream architecture while deeply exploiting generative diffusion priors. Rather than solely utilizing diffusion models as monolithic augmentors via LoRA-guided synthesis to generate paired training data, we innovatively repurpose a pre-trained diffusion-based geometric estimator as a non-linear low-pass filter. This mathematically establishes an estimated stream comprising purely low-frequency information, which serves as an auxiliary anchor for the stable macroscopic structural representation, constructing a robust nominal template even with single-digit samples. Concurrently, the uncompressed real stream, encompassing coupled high- and low-frequency components, is explicitly enhanced by this auxiliary prior to capture localized microscopic details. By synergistically aggregating the anomaly responses from these parallel streams, our framework naturally isolates and amplifies genuine defect signals while effectively suppressing cross-modal interference.

To precisely align these decoupled dual-stream representations without risking feature collapse caused by cross-modal amplitude discrepancies, we design an adaptive Coordinate-Aware Hierarchical Feature Mapper. Unlike previous rigid fusion mechanisms, it dynamically regulates hierarchical feature aggregation within each independent mapping pathway across multiple scales, while rigorously preserving spatial positional priors. Finally, we introduce a Cross-Modal Multiplicative Anomaly Scoring module ($\Psi_{\text{2D}} \odot \Psi_{\text{3D}}$). This acts as a stringent spatial filter that flags defects only given concurrent multi-modal anomalies, dramatically suppressing previously unavoidable modality-specific false alarms triggered by localized sensory artifacts.

Our main contributions are summarized as follows:
\begin{enumerate}
    \item \textbf{Cross-Modal Dual-Stream Framework (CMDS-AD):} We propose the first few-shot MAD architecture driven by a dual-stream perspective. By innovatively repurposing diffusion estimators as non-linear low-pass filters, we establish an auxiliary estimated stream that effectively enhances the real stream to isolate unpredictable micro-defects from stable macroscopic structures.
    \item \textbf{Adaptive Cross-Modal Feature Alignment:} We design a Coordinate-Aware Hierarchical Feature Mapper coupled with a Decoupled Multi-Scale Mask-Aware Optimization strategy. This adaptively aligns heterogeneous 2D and 3D semantic spaces, fundamentally mitigating the cross-modal gap and modality interference under severe data scarcity.
    \item \textbf{Synergistic Multiplicative Anomaly Scoring:} We introduce a novel scoring mechanism that aggregates anomaly responses from both the real and estimated streams to filter out isolated modality-specific noise. Extensive experiments on MVTec 3D-AD and EyeCandies demonstrate that our method establishes a new state-of-the-art in few-shot settings.
\end{enumerate}

\section{Related Work}
\label{sec:related}

\noindent\textbf{Multi-modal Anomaly Detection.} 
With the advent of 3D sensors, multi-modal anomaly detection (MAD) has gained significant attention \cite{wang2023multimodal, chu2023shape, costanzino2024multimodal} as 3D geometric cues effectively complement RGB features. Recent methods typically explore cross-modal feature alignment (\eg CFM \cite{costanzino2024multimodal}, CIF \cite{lin2026commonality}), memory-based architectures (\eg ShapeGuided \cite{chu2023shape}, M3DM \cite{wang2023multimodal}, MAFR \cite{ali20252d}), or multi-level feature fusion (\eg MMRD \cite{gu2024rethinking}) to localize anomalies. Although these approaches achieve remarkable performance under full-shot settings, they heavily rely on extensive normal training data or perfectly aligned networks. In severe few-shot scenarios, the profound modality gap between 2D images and 3D point clouds critically hinders reliable cross-modal alignment. To address this fundamental limitation, we propose a Coordinate-Aware Hierarchical Feature Mapper that bypasses amplitude discrepancies and adaptively aligns multi-scale contextual features, efficiently closing the modality gap under data scarcity.

\noindent\textbf{Few-shot Anomaly Detection.} 
To alleviate the prohibitive cost of collecting large-scale normal data, few-shot anomaly detection (FSAD) has emerged as a promising alternative. Current research focuses on the RGB domain, utilizing vision-language prompts (AnoPLe \cite{lee2024anople}), feature regression (FastRecon \cite{fang2023fastrecon}), window-based fusion (WinCLIP \cite{jeong2023winclip}), or image registration (RegAD \cite{huang2022registration}). However, existing single-modality FSAD approaches struggle to capture structural distortions. Extending FSAD to multi-modal scenarios remains challenging because traditional methods uniformly treat all spatial features, exacerbating modality interference when data is scarce. Motivated by this, our method explicitly establishes a dual-stream architecture to process coupled and purely low-frequency representations, effectively mitigating cross-modal interference.

\noindent\textbf{Diffusion Models in Visual Inspection.} 
Diffusion models are increasingly applied in anomaly detection for data augmentation \cite{trabucco2024effective, lyu2024reb} or reconstruction \cite{wyatt2022anoddpm, zhang2025diffusionad, he2024diffusion}. Restoration studies further show that adaptive denoising schedules and lightweight LoRA adaptation can target image degradations, as in SkipDiff \cite{luo2024skipdiff} and degradation-aware LoRA \cite{tang2026diffusion}. These advances suggest that diffusion priors can be adapted beyond generic generation, but their frequency-level role in multi-modal inspection remains underexplored. However, existing anomaly detection methods still treat generation as a monolithic module, overlooking the frequency dynamics of progressive denoising \cite{choi2022perception}. Distinctly, our framework rethinks their role in FSAD. While utilizing LoRA \cite{hu2022lora} for paired data synthesis to mitigate sample scarcity, we further draw on recent findings demonstrating that diffusion estimators inherently act as non-linear low-pass filters \cite{si2024freeu, yang2023diffusion}. Leveraging this property, we establish an auxiliary estimated stream containing purely low-frequency information to anchor macroscopic structures, which enhances the sensitive real stream containing coupled high- and low-frequency information to capture micro-defects. This explicitly bridges generative priors with dual-stream alignment, resolving the overfitting bottlenecks of few-shot generation.

\section{Methodology}

\subsection{Overall Pipeline Architecture}
\begin{figure*}[t!]
    \centering
    \includegraphics[width=1.0\textwidth]{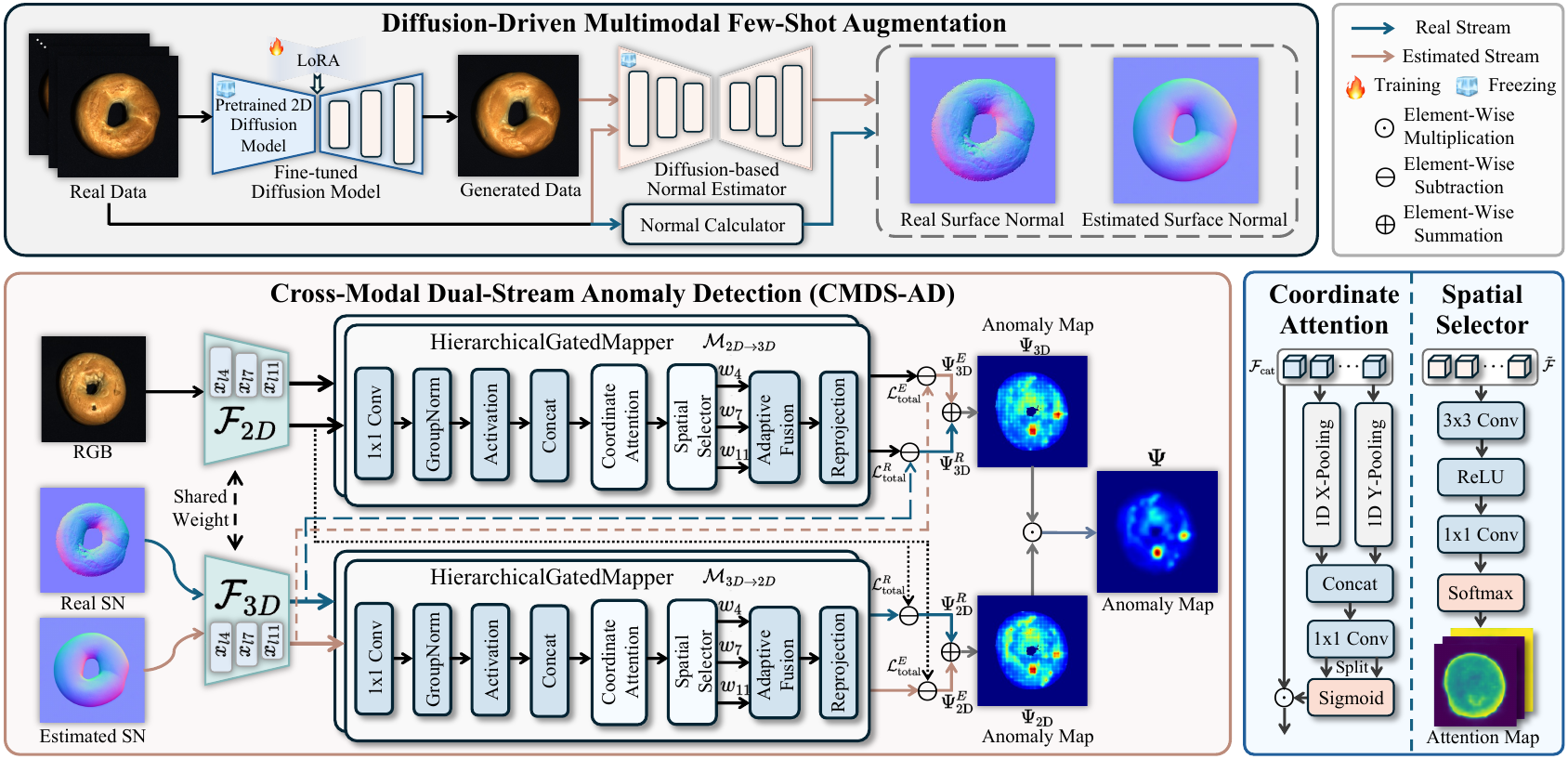}
    \caption{Proposed Cross-Modal Dual-Stream Anomaly Detection (CMDS-AD) framework. \textbf{(Top)} Diffusion-driven augmentation synthesizes paired RGB-normal training data. \textbf{(Bottom Left)} Dual-stream pipeline processing Real and Estimated streams, where the purely low-frequency estimated stream acts as an auxiliary guide to enhance the coupled real stream. \textbf{(Bottom Right)} Coordinate-Aware Hierarchical Feature Mapper for adaptive multi-scale feature fusion.}
    \label{fig:pipeline}
\end{figure*}

Figure \ref{fig:pipeline} shows our CMDS-AD framework, designed to bridge the 2D-3D semantic gap under few-shot constraints. To overcome data scarcity, a Diffusion-Driven Multimodal Few-Shot Augmentation module (Sec. \ref{sec:diffusion_aug}) first synthesizes paired RGB-normal training data. The pipeline then employs a bidirectional mapping mechanism (2D $\leftrightarrow$ 3D) comprising parallel real and estimated streams. Given the input modalities, a frozen backbone (e.g., ViT) extracts multi-scale features, which a Coordinate-Aware Hierarchical Feature Mapper (Sec. \ref{sec:feature_gating}) adaptively aggregates via spatial gating. During training, a Decoupled Multi-Scale Mask-Aware Optimization strategy (Sec. \ref{sec:optimization}) ensures precise alignment across the heterogeneous domains. During inference, the mappers ($\mathcal{M}_{\text{2D} \to \text{3D}}, \mathcal{M}_{\text{3D} \to \text{2D}}$) generate directional anomaly distance maps for both streams. Finally, a Cross-Modal Multiplicative Anomaly Scoring module (Sec. \ref{sec:scoring}) synergistically fuses these directional predictions into a high-precision dense anomaly map $\Psi$, thereby effectively filtering out modality-specific artifacts to reduce false positives. The complete procedure is summarized in Algorithm \ref{alg:fsad_pipeline}.

% ==================================== Pipeline =====================================

\subsection{Diffusion-Driven Multimodal Few-Shot Augmentation}
\label{sec:diffusion_aug}
In Few-Shot Anomaly Detection (FSAD), extreme data scarcity severely bottlenecks model generalization. To overcome this, we propose a multimodal data augmentation framework driven by diffusion priors. For the 2D (RGB) modality, we adopt a Low-Rank Adaptation (LoRA)-guided image-to-image generation paradigm. Standard diffusion models learn data distributions by optimizing the denoising objective:
\begin{equation}
\label{eq:dm_loss}
\mathcal{L}_{\text{DM}} = \mathbb{E}_{x_0, \epsilon \sim \mathcal{N}(0,\mathbf{I}), t} \left[ \|\epsilon - \epsilon_\theta(x_t, t)\|_2^2 \right]
\end{equation}
where $x_0$ is the clean image, $\epsilon$ represents the injected Gaussian noise, $t$ is the timestep, and $\epsilon_\theta$ denotes the denoising network parameterized by $\theta$. Directly fine-tuning $\theta$ on limited samples inevitably triggers catastrophic forgetting and severe overfitting. Thus, we freeze the pre-trained projection matrices $W_0 \in \mathbb{R}^{d \times k}$ in the attention layers and inject trainable low-rank matrices $B \in \mathbb{R}^{d \times r}$ and $A \in \mathbb{R}^{r \times k}$ (with $r \ll \min(d,k)$). The forward pass update is formulated as:
\begin{equation}
\label{eq:lora_weight}
W = W_0 + \Delta W = W_0 + BA
\end{equation}
where $W$ is the updated weight matrix and $\Delta W$ is the weight increment. This parameter-efficient formulation enables the model to extract domain-specific textural features from scarce normal samples, synthesizing structurally consistent yet distributionally diverse RGB representations. For the 3D modality, surface normal maps are more sensitive to subtle geometric discontinuities (\eg scratches, dents) than depth maps. Consequently, we introduce a diffusion-based normal estimator. By processing both the real and LoRA-generated RGB images through this estimator, we obtain their estimated normal maps, completing the real-estimated cross-modal augmented dataset.

\subsection{Dual-Stream Decoupled Anomaly Detection}
During the pipeline, the network jointly processes the real surface normals and the estimated normals generated by the diffusion estimator $\Phi_{\text{diff}}$. Constrained by generative priors and latent space compression, $\Phi_{\text{diff}}$ acts essentially as a robust non-linear low-pass filter, tending to output overly smooth results that inadvertently discard high-frequency details.

To formalize this theoretically, in the frequency domain, the real surface normal $N$ encompasses a low-frequency structural component $N_{\text{low}}$ (global shape) and a high-frequency detail component $N_{\text{high}}$ (local micro-textures). For an input image $I$, the estimator yields a smoothed estimation $\hat{N} = \Phi_{\text{diff}}(I) \approx N_{\text{low}}$. Consequently, mapping these inputs through the backbone $\Phi$ yields two complementary feature spaces rather than a single entangled representation:
\begin{equation}
\label{eq:frequency_decoupling}
\mathcal{F}_{\text{est}} \approx \Phi(N_{\text{low}}), \quad \mathcal{F}_{\text{real}} \approx \Phi(N_{\text{low}} + N_{\text{high}})
\end{equation}

Leveraging this inherent filtering property, we design a synergistic dual-stream architecture. To avoid amplifying cross-modal noise via explicit feature residuals, we process both streams independently. The estimated stream ($\mathcal{F}_{\text{est}}$) acts as a purely low-frequency auxiliary anchor for macroscopic structures. Guided by this reference, the uncompressed real stream ($\mathcal{F}_{\text{real}}$)—which couples high- and low-frequency components—precisely captures localized surface defects. Synergizing the independent anomaly measurements from these distinct sub-spaces ensures comprehensive defect coverage.

% ================================ Algorithm ========================================
\begin{algorithm}[t!]
\small
\LinesNumbered
\SetAlgoLined
\KwIn{Support set $\mathcal{D}_{\text{train}} = \{I_{\text{rgb}}, N, M\}$, Pre-trained backbone $\Phi$, Untrained Mappers $\mathcal{M}_{\text{2D} \to \text{3D}}, \mathcal{M}_{\text{3D} \to \text{2D}}$}
\KwOut{Optimized Mappers $\mathcal{M}_{\text{2D} \to \text{3D}}, \mathcal{M}_{\text{3D} \to \text{2D}}$, Final Anomaly Map $\Psi$}
\BlankLine

\textbf{Stage 1: Multimodal Data Augmentation}\\
$\{A, B\} \leftarrow \arg\min_{A, B} \mathcal{L}_{\text{DM}}$ \tcp*{LoRA FT via Eq. \ref{eq:dm_loss} and \ref{eq:lora_weight}}
$\hat{I}_{\text{rgb}} \sim \text{DM}_{W_0+BA}(I_{\text{rgb}}), \quad \hat{N} \leftarrow \Phi_{\text{diff}}(\hat{I}_{\text{rgb}})$ \tcp*{Generate cross-modal pairs}
$\mathcal{D}_{\text{aug}} \leftarrow \mathcal{D}_{\text{train}} \cup \{(\hat{I}_{\text{rgb}}, \hat{N})\}$\;

\BlankLine
\textbf{Stage 2: Hierarchical Feature Mapper Optimization}\\
\For{batch $\mathcal{B} \sim \mathcal{D}_{\text{aug}}$}{
    $\mathcal{F}_{\text{2D}}^{S}, \mathcal{F}_{\text{3D}}^{S} \leftarrow \Phi(\mathcal{B}), \quad S \in \{R, E\}$ \tcp*{Extract $l \in \{4,7,11\}$ layers}
    
    \tcp{Coordinate-Aware Mapping (Sec. \ref{sec:feature_gating})}
    $\mathcal{P}_{\text{3D}}^{S} \leftarrow \mathcal{M}_{\text{2D} \to \text{3D}}(\mathcal{F}_{\text{2D}}^{S}), \quad S \in \{R, E\}$ \tcp*{2D $\to$ 3D direction}
    $\mathcal{P}_{\text{2D}}^{S} \leftarrow \mathcal{M}_{\text{3D} \to \text{2D}}(\mathcal{F}_{\text{3D}}^{S}), \quad S \in \{R, E\}$ \tcp*{3D $\to$ 2D direction}
    
    \tcp{Decoupled Mask-Aware Optimization (Sec. \ref{sec:optimization})}
    \For{layer $l \in \{4, 7, 11\}$}{
        $\mathcal{L}_{\text{L}l}^{R} \leftarrow \text{Eq. \ref{eq:mask_loss}}\big(\mathcal{L}_{\text{align}}(\mathcal{P}_{\text{L}l}^{R}, \mathcal{F}_{\text{L}l}^{R}), M\big)$ \tcp*{Masked avg for Real}
        $\mathcal{L}_{\text{L}l}^{E} \leftarrow \text{Eq. \ref{eq:mask_loss}}\big(\mathcal{L}_{\text{align}}(\mathcal{P}_{\text{L}l}^{E}, \mathcal{F}_{\text{L}l}^{E})\big)$ \tcp*{Global avg for Est}
    }
    $\mathcal{L}_{\text{total}}^{S} \leftarrow \alpha \mathcal{L}_{\text{L4}}^{S} + \beta \mathcal{L}_{\text{L7}}^{S} + \gamma \mathcal{L}_{\text{L11}}^{S}, \quad S \in \{R, E\}$ \tcp*{Eq. \ref{eq:total_loss}}
    $\mathcal{L} \leftarrow \mathcal{L}_{\text{total}}^{R} + \mathcal{L}_{\text{total}}^{E}$ \tcp*{Overall objective}
    $\{\mathcal{M}_{\text{2D} \to \text{3D}}, \mathcal{M}_{\text{3D} \to \text{2D}}\} \xleftarrow{\nabla} \mathcal{L}$ \tcp*{Backpropagation}
}

\BlankLine
\textbf{Stage 3: Cross-Modal Multiplicative Anomaly Scoring}\\
$\mathcal{F}_{\text{2D}}^{S}, \mathcal{F}_{\text{3D}}^{S} \leftarrow \Phi(X_{\text{test}}), \quad S \in \{R, E\}$\;
$\mathcal{P}_{\text{3D}}^{S} \leftarrow \mathcal{M}_{\text{2D} \to \text{3D}}(\mathcal{F}_{\text{2D}}^{S}), \quad \mathcal{P}_{\text{2D}}^{S} \leftarrow \mathcal{M}_{\text{3D} \to \text{2D}}(\mathcal{F}_{\text{3D}}^{S})$\;
$\Psi_{\text{2D}}^{S} \leftarrow \mathcal{L}_{\text{align}}(\mathcal{P}_{\text{2D}}^{S}, \mathcal{F}_{\text{2D}}^{S}), \quad \Psi_{\text{3D}}^{S} \leftarrow \mathcal{L}_{\text{align}}(\mathcal{P}_{\text{3D}}^{S}, \mathcal{F}_{\text{3D}}^{S})$ \tcp*{Eq. \ref{eq:cosine_loss}}
$\Psi_{\text{2D}} \leftarrow \Psi_{\text{2D}}^{R} + \lambda_1 \Psi_{\text{2D}}^{E}, \quad \Psi_{\text{3D}} \leftarrow \Psi_{\text{3D}}^{R} + \lambda_2 \Psi_{\text{3D}}^{E}$ \tcp*{Eq. \ref{eq:modality_anomaly}}
$\Psi \leftarrow \Psi_{\text{2D}} \odot \Psi_{\text{3D}}$ \tcp*{Eq. \ref{eq:anomaly_fusion}}
\KwRet{$\Psi$}
\caption{CMDS-AD: Cross-Modal Dual-Stream Anomaly Detection}
\label{alg:fsad_pipeline}
\end{algorithm}

\subsection{Coordinate-Aware Hierarchical Feature Mapper}
\label{sec:feature_gating}
To align multimodal features across varying receptive fields, we design a feature mapper comprising hierarchical branches. The backbone (\eg ViT) extracts multi-scale representations from the 4-th, 7-th, and 11-th layers, denoted as $x_{l} \in \mathbb{R}^{C \times H \times W}$ for $l \in \{4, 7, 11\}$ (where $C=768$). These represent local edge textures, part-level patterns, and global semantics, respectively. Prior to spatial gating, each extracted feature $x_l$ first passes through a local adaptation block—comprising a $1 \times 1$ convolution, Group Normalization (GN), and a GELU activation—to harmonize the heterogeneous semantic spaces across different network depths. The adapted features are subsequently concatenated along the channel dimension to form a unified multi-scale input feature $\mathcal{F}_{\text{cat}} \in \mathbb{R}^{3C \times H \times W}$:
\begin{equation}
\label{eq:feature_concat}
\hat{x}_l = \text{GELU}\big(\text{GN}(\text{Conv}_{1 \times 1}(x_l))\big), \quad \mathcal{F}_{\text{cat}} = \text{Concat}(\hat{x}_4, \hat{x}_7, \hat{x}_{11})
\end{equation}

Since anomaly detection requires spatially sensitive dense predictions, traditional Channel Attention is sub-optimal as it collapses spatial dimensions via Global Average Pooling. Instead, we introduce Coordinate Attention (CA), which factorizes 2D spatial pooling into two 1D directional encoding operations. For the $c$-th channel of $\mathcal{F}_{\text{cat}}$ at coordinates $(h, w)$, the aggregated features along height and width are computed as:
\begin{equation}
\label{eq:coordinate_attention}
z_c^h(h) = \frac{1}{W} \sum_{i=0}^{W-1} \mathcal{F}_{\text{cat}}^{(c)}(h,i), \quad z_c^w(w) = \frac{1}{H} \sum_{j=0}^{H-1} \mathcal{F}_{\text{cat}}^{(c)}(j,w)
\end{equation}

These direction-aware vectors are concatenated, transformed, and split to generate attention weights, achieving feature refinement without sacrificing positional precision. The refined concatenated feature $\tilde{\mathcal{F}}$ is then fed into a lightweight Spatial Selector. This module generates a mutually exclusive 3D weight map $\mathcal{W} \in \mathbb{R}^{3 \times H \times W}$ (where $\mathcal{W} = [\omega_4, \omega_7, \omega_{11}]$) via a pixel-wise Softmax operation:
\begin{equation}
\label{eq:spatial_selector}
[\omega_4, \omega_7, \omega_{11}] = \text{Softmax}\big(\text{Conv}_{1 \times 1}(\text{ReLU}(\text{Conv}_{3 \times 3}(\tilde{\mathcal{F}})))\big)
\end{equation}
where $\omega_l$ indicates the dynamic aggregation weight of the $l$-th layer at a specific pixel ($\sum_l \omega_l = 1$). The preprocessed single-layer features $f_l$ are then spatially fused and projected to yield the predicted feature $\mathcal{P} = \Pi\!\left(\sum_{l \in \{4,7,11\}} \omega_l \odot f_l\right)$.

\subsection{Decoupled Multi-Scale Mask-Aware Optimization}
\label{sec:optimization}
To compel the mapping network to finely align local microscopic textures while robustly matching global macroscopic structures, we propose a Decoupled Multi-Scale Loss governed by a divide-and-conquer strategy.

\noindent \textbf{Feature Decoupling and Alignment.} Prior to loss computation, the unified predicted feature $\mathcal{P}$ is decoupled back into three distinct representation sub-spaces along the multi-scale spectrum (ranging from high-frequency local textures to low-frequency global structures, each with 768 dimensions). Given the massive absolute numerical discrepancies between RGB and 3D normal domains, we discard magnitude-sensitive distance metrics and rely exclusively on Cosine Distance as the core alignment objective:
\begin{equation}
\label{eq:cosine_loss}
\mathcal{L}_{\text{align}}(P, T) = 1 - \frac{P \cdot T}{\|P\|_2 \|T\|_2}
\end{equation}
where $P, T \in \mathbb{R}^{768}$ denote the predicted and target feature vectors at a specific pixel. This compels the network to align strictly based on directional consistency, circumventing gradient collapse caused by cross-modal amplitude variations.

\noindent \textbf{Mask-Aware Regional Focusing.} To prevent massive irrelevant backgrounds (\eg conveyors) from dominating the loss, we introduce a flattened binary spatial mask $M \in \{0,1\}^{N}$ (where $N = H \times W$). For the real stream, adaptive focusing is achieved by computing the masked average ($\mathcal{L}_{\text{layer}}^{R}$). Conversely, for the estimated stream where precise masks are unavailable, we default to global spatial averaging ($\mathcal{L}_{\text{layer}}^{E}$):
\begin{equation}
\label{eq:mask_loss}
\mathcal{L}_{\text{layer}}^{R} = \frac{\sum_{i=1}^{N} \left(\mathcal{L}_{\text{align}}^{(i)} \cdot M^{(i)}\right)}{\sum_{i=1}^{N} M^{(i)} + \epsilon}, \quad \mathcal{L}_{\text{layer}}^{E} = \frac{1}{N} \sum_{i=1}^{N} \mathcal{L}_{\text{align}}^{(i)}
\end{equation}
where $\epsilon = 10^{-8}$, superscript $(i)$ denotes the $i$-th pixel index, and the binary mask $M$ is deterministically computed from the raw 3D point cloud to precisely isolate the foreground object.

\noindent \textbf{Hierarchical Weighted Fusion.} Deep features exhibit high spatial translation invariance; enforcing strict pixel-wise alignment at semantic levels easily triggers overfitting. Thus, for both the real ($R$) and estimated ($E$) streams, we introduce an asymmetric descending penalty strategy based on their respective layer-wise losses:
\begin{equation}
\label{eq:total_loss}
\mathcal{L}_{\text{total}}^{S} = \alpha \mathcal{L}_{\text{L4}}^{S} + \beta \mathcal{L}_{\text{L7}}^{S} + \gamma \mathcal{L}_{\text{L11}}^{S}, \quad S \in \{R, E\}
\end{equation}
where weights satisfy $\alpha > \beta > \gamma$. This strategically anchors macroscopic structural comprehension (via a smaller $\gamma$ for deep layers) while intensely directing focal attention towards critical localized micro-defects (via a larger $\alpha$ for shallow layers). The overall optimization objective of our framework is the sum of both streams: $\mathcal{L} = \mathcal{L}_{\text{total}}^{R} + \mathcal{L}_{\text{total}}^{E}$.

\subsection{Cross-Modal Multiplicative Anomaly Scoring}
\label{sec:scoring}
During inference, the framework yields four distinct anomaly distance maps: 2D real $\Psi_{\text{2D}}^{R}$, 2D estimated $\Psi_{\text{2D}}^{E}$, 3D real $\Psi_{\text{3D}}^{R}$, and 3D estimated $\Psi_{\text{3D}}^{E}$. To obtain comprehensive modality-specific responses, we first perform a weighted summation of the real and estimated streams within both the 2D and 3D modalities:
\begin{equation}
\label{eq:modality_anomaly}
\Psi_{\text{2D}} = \Psi_{\text{2D}}^{R} + \lambda_1 \Psi_{\text{2D}}^{E}, \quad \Psi_{\text{3D}} = \Psi_{\text{3D}}^{R} + \lambda_2 \Psi_{\text{3D}}^{E}
\end{equation}
where $\lambda_1$ and $\lambda_2$ are balancing coefficients. Subsequently, the overall anomaly score map $\Psi$ is computed via synergistic cross-modal fusion:
\begin{equation}
\label{eq:anomaly_fusion}
\Psi = \Psi_{\text{2D}} \odot \Psi_{\text{3D}}
\end{equation}
where $\odot$ denotes the Hadamard product. This multiplicative fusion acts as a stringent spatial filter: a region is flagged as defective only if both modalities exhibit high anomalous responses concurrently, thereby significantly suppressing isolated modality-specific noise and minimizing false positives.

\section{Experiments}

\subsection{Implementation Setup}
\noindent \textbf{Datasets and Protocol.} 
We evaluate our framework on two standard anomaly detection datasets: MVTec 3D-AD \cite{bergmann2021mvtec} and EyeCandies \cite{bonfiglioli2022eyecandies}. Following the standard few-shot protocol, models are trained exclusively on $k \in \{1, 2, 4\}$ anomaly-free samples per category. To ensure a standardized and fully reproducible evaluation, we deterministically select the first $k$ normal images from each class to construct the training subset.

\noindent \textbf{Baselines.} 
We compare our approach against eight representative state-of-the-art methods: (1) \textbf{Patchcore+FPFH} \cite{horwitz2023back}, concatenating deep 2D and hand-crafted 3D features; (2) \textbf{CIF} \cite{lin2026commonality}, a cross-modal information fusion strategy; (3) \textbf{AST} \cite{rudolph2023asymmetric}, an asymmetric teacher-student architecture; (4) \textbf{EasyNet} \cite{chen2023easynet}, an efficient network tailored for industrial inspection; (5) \textbf{ShapeGuided} \cite{chu2023shape}, explicitly leveraging 3D shape priors; (6) \textbf{M3DM} \cite{wang2023multimodal}, a memory-bank-based multi-modal fusion model; (7) \textbf{CFM} \cite{costanzino2024multimodal}, a memory-efficient 2D-3D feature alignment approach; and (8) \textbf{MAFR} \cite{ali20252d}, a multi-modal feature representation and alignment network.

\noindent \textbf{Evaluation Metrics.} 
Following standard protocols, we employ Image-level AUROC (I-AUROC) for anomaly classification, alongside Pixel-level AUROC (P-AUROC) and Area Under Per-Region Overlap (AUPRO) for fine-grained localization. In our main baseline comparisons, we report I-AUROC and AUPRO integrated up to a 30\% False Positive Rate (FPR). For a comprehensive analysis in the ablation studies, we additionally report P-AUROC and AUPRO evaluated at stricter FPR thresholds of 1\%, 5\%, and 10\%. Higher scores across all metrics denote superior performance.

\noindent \textbf{Implementation Details.} 
Experiments utilize an RTX 5090 (32GB) GPU. For augmentation, Stable Diffusion v2.1 is LoRA-finetuned (rank $16$, $1{,}000$ steps), and Marigold \cite{ke2025marigold} serves as the normal estimator $\Phi_{\text{diff}}$. Following \cite{costanzino2024multimodal}, a frozen DINO ViT-B/8 extracts 2D and 3D representations \cite{li2025find} from $224 \times 224$ inputs. Features from layers 4, 7, and 11 are bicubically up-sampled and optimized via AdamW for $3{,}000$ steps (batch size $2$/stream, Cosine Annealing LR $10^{-4}$ to $10^{-6}$, weight decay $10^{-4}$). Additive Gaussian noise ($\sigma = 0.01$) is injected into RGB inputs to mitigate overfitting. Multi-scale loss weights (Eq. \ref{eq:total_loss}) are set to $\alpha=1.2, \beta=1.0, \gamma=0.8$. Inference balancing coefficients ($\lambda_1, \lambda_2$) are $0.1$.

%===============================MVTec 3D-AD Result Updated==============================
\begin{table*}[t!]
\centering
\caption{Few-shot performance comparison on the MVTec 3D-AD dataset. The metrics reported are I-AUROC / AUPRO@30\%. The best results are highlighted in \textbf{bold}, and the second-best are \underline{underlined}.}
\label{tab:few_shot_combined_updated}
\resizebox{\textwidth}{!}{
\renewcommand\arraystretch{1.1}
\begin{tabular}{c|c|l ccccccccccc}
\toprule
\rowcolor{gray!20}
Setting & Type & Method & Bagel & Cable Gland & Carrot & Cookie & Dowel & Foam & Peach & Potato & Rope & Tire & Mean \\
\midrule
\multirow{9}{*}{\rotatebox{90}{1-shot}} 
& \multirow{2}{*}{TF} 
  & Patchcore+FPFH~\cite{horwitz2023back}    & 62.2/92.8 & 53.4/76.8 & 54.0/96.7 & 55.9/92.8 & 54.7/84.6 & 63.3/71.9 & 49.6/95.9 & 60.7/96.1 & 88.8/90.8 & 56.6/84.1 & 59.9/88.3 \\
& & CIF~\cite{lin2026commonality}               & 78.9/85.2 & \textbf{68.7}/79.5 & 72.2/95.6 & 81.2/86.2 & \underline{62.9}/82.0 & \textbf{72.8}/70.3 & 83.9/93.3 & 60.4/94.2 & 83.5/88.1 & 55.8/86.4 & 72.0/86.1 \\
\cmidrule{2-14}
& \multirow{7}{*}{TB} 
  & AST~\cite{rudolph2023asymmetric}               & 70.7/75.9 & 42.2/73.3 & 54.8/88.0 & 49.0/60.2 & 53.8/79.4 & 46.4/44.0 & 51.9/84.0 & 49.7/85.9 & 72.0/75.8 & 41.9/74.0 & 53.2/74.0 \\
& & EasyNet~\cite{chen2023easynet}           & 61.4/79.6 & 21.2/75.1 & 52.0/91.0 & 75.9/69.8 & 56.5/85.8 & 62.8/49.4 & 65.7/69.0 & 63.0/88.1 & 94.6/71.8 & 47.7/75.4 & 60.1/75.5 \\
& & ShapeGuided~\cite{chu2023shape}       & 65.9/\underline{95.9} & 44.4/71.6 & 62.3/93.5 & 93.8/\textbf{94.1} & 59.3/86.4 & 57.6/63.8 & 67.6/94.0 & 42.8/96.3 & 93.3/88.8 & \underline{62.9}/90.1 & 65.0/87.4 \\
& & M3DM~\cite{wang2023multimodal}              & 87.8/95.3 & \underline{64.1}/\underline{81.5} & \underline{78.0}/\underline{97.2} & 92.7/90.3 & \textbf{64.2}/81.6 & 65.3/82.0 & 75.5/94.0 & \textbf{79.8}/94.8 & 85.8/95.2 & 45.4/89.8 & \underline{73.9}/90.2 \\
& & CFM~\cite{costanzino2024multimodal}               & \underline{88.1}/95.4 & 54.0/74.0 & 62.5/96.5 & \underline{96.4}/92.4 & 60.2/\textbf{91.4} & 68.7/89.3 & 61.9/93.8 & 56.0/95.6 & 90.9/95.7 & 58.8/88.9 & 69.8/91.4 \\
& & MAFR~\cite{ali20252d}              & 86.7/95.0 & 48.9/78.3 & 70.1/96.9 & \textbf{96.7}/92.2 & 56.9/88.0 & 59.9/\textbf{89.7} & \underline{78.4}/\underline{96.6} & 67.3/\underline{96.8} & \underline{95.3}/\underline{96.8} & \textbf{64.0}/\underline{91.8} & 72.4/\underline{92.2} \\
\rowcolor{cvprblue!15} 
\cellcolor{white} & \cellcolor{white} & \textit{\textbf{Ours}}     & \textbf{97.0}/\textbf{97.0} & 63.8/\textbf{88.0} & \textbf{92.0}/\textbf{98.1} & 92.3/\underline{93.2} & 59.6/\underline{88.5} & \underline{69.9}/\underline{89.4} & \textbf{85.6}/\textbf{97.9} & \underline{77.4}/\textbf{98.0} & \textbf{97.1}/\textbf{97.6} & 61.4/\textbf{93.8} & \textbf{79.6}/\textbf{94.2} \\

\midrule
\multirow{9}{*}{\rotatebox{90}{2-shot}} 
& \multirow{2}{*}{TF} 
  & Patchcore+FPFH~\cite{horwitz2023back}    & 63.2/94.6 & 47.7/76.5 & 55.4/96.7 & 64.5/\underline{93.4} & 58.3/85.0 & 61.1/67.4 & 55.0/96.2 & 56.6/96.6 & 88.2/91.0 & 64.4/88.3 & 61.4/88.6 \\
& & CIF~\cite{lin2026commonality}               & 85.3/87.1 & \underline{62.9}/79.5 & 74.0/95.8 & 72.2/87.1 & 64.6/82.1 & \underline{79.5}/75.3 & 77.6/93.7 & 66.9/94.9 & 86.1/88.7 & 62.7/87.7 & 73.2/87.2 \\
\cmidrule{2-14}
& \multirow{7}{*}{TB} 
  & AST~\cite{rudolph2023asymmetric}               & 71.9/75.9 & 43.4/74.0 & 54.5/87.8 & 50.8/62.2 & 53.7/79.5 & 46.1/43.6 & 51.6/83.7 & 50.4/85.6 & 75.8/76.5 & 40.2/72.8 & 53.8/74.2 \\
& & EasyNet~\cite{chen2023easynet}           & 47.6/77.8 & \textbf{76.1}/62.9 & 52.6/92.6 & 60.2/59.1 & 31.7/58.8 & 52.3/57.2 & 71.9/21.1 & \underline{76.1}/15.2 & 61.2/43.1 & 51.2/4.7 & 58.1/49.3 \\
& & ShapeGuided~\cite{chu2023shape}       & 47.9/\underline{96.6} & 46.0/73.2 & 60.5/96.5 & \underline{95.9}/\textbf{95.5} & 55.3/86.5 & 50.2/71.4 & 69.7/95.3 & 41.3/96.3 & 93.6/89.3 & \textbf{79.3}/91.3 & 64.0/89.2 \\
& & M3DM~\cite{wang2023multimodal}              & \underline{91.8}/95.5 & 57.0/\underline{82.9} & \underline{79.8}/97.2 & 94.5/88.0 & 61.4/87.0 & \underline{79.5}/79.6 & 79.2/95.1 & 75.1/94.2 & 92.8/95.5 & 54.1/91.0 & 76.5/90.6 \\
& & CFM~\cite{costanzino2024multimodal}               & 88.9/95.9 & 53.3/76.2 & 69.6/\underline{97.7} & 94.5/92.5 & \underline{70.1}/\textbf{93.0} & 77.6/\textbf{92.6} & 63.9/94.2 & 63.8/\underline{97.3} & 91.5/96.3 & 57.2/90.7 & 73.0/92.4 \\
& & MAFR~\cite{ali20252d}              & \underline{91.8}/95.7 & 49.8/79.7 & 69.8/97.1 & \textbf{96.9}/92.0 & \textbf{70.5}/\underline{89.8} & 71.4/\underline{92.2} & \underline{81.2}/\underline{97.0} & 72.8/97.1 & \underline{95.3}/\underline{96.9} & 66.9/\underline{94.2} & \underline{76.6}/\underline{93.2} \\
\rowcolor{cvprblue!15} 
\cellcolor{white} & \cellcolor{white} & \textit{\textbf{Ours}}     & \textbf{96.4}/\textbf{97.2} & 62.3/\textbf{89.1} & \textbf{93.0}/\textbf{98.2} & 93.5/93.3 & 58.8/89.3 & \textbf{87.7}/\textbf{92.6} & \textbf{86.3}/\textbf{97.8} & \textbf{83.5}/\textbf{98.1} & \textbf{97.6}/\textbf{97.7} & \underline{70.7}/\textbf{94.8} & \textbf{83.0}/\textbf{94.8} \\

\midrule
\multirow{9}{*}{\rotatebox{90}{4-shot}} 
& \multirow{2}{*}{TF} 
  & Patchcore+FPFH~\cite{horwitz2023back}    & 52.3/95.7 & 54.0/79.7 & 59.2/97.2 & 62.9/\underline{95.1} & 59.4/87.5 & 58.2/75.3 & 61.6/96.5 & 70.5/97.3 & 91.8/91.2 & 73.5/88.0 & 64.3/90.4 \\
& & CIF~\cite{lin2026commonality}               & 91.8/92.9 & \underline{70.0}/83.3 & 77.5/96.9 & 83.4/86.4 & 70.2/84.7 & 79.4/84.1 & \underline{85.6}/95.2 & 75.0/95.5 & 89.8/89.3 & 53.4/88.2 & 77.6/89.6 \\
\cmidrule{2-14}
& \multirow{7}{*}{TB} 
  & AST~\cite{rudolph2023asymmetric}               & 70.1/74.7 & 42.9/73.7 & 55.7/87.7 & 51.8/61.3 & 54.0/79.6 & 46.6/41.3 & 52.0/84.3 & 49.8/85.9 & 72.6/75.9 & 39.8/74.2 & 53.5/73.9 \\
& & EasyNet~\cite{chen2023easynet}           & 67.5/70.7 & 36.3/13.8 & 54.7/86.9 & 69.1/72.0 & 72.4/39.3 & 50.2/53.6 & 74.3/86.2 & 63.1/90.2 & 44.9/15.5 & 53.7/66.3 & 58.6/56.5 \\
& & ShapeGuided~\cite{chu2023shape}       & 65.4/\textbf{97.3} & 48.8/78.9 & 73.1/97.3 & 96.5/\textbf{95.4} & 69.8/90.4 & 59.1/83.6 & 68.1/95.7 & 49.9/97.5 & 92.2/89.6 & \underline{75.1}/92.1 & 69.8/91.8 \\
& & M3DM~\cite{wang2023multimodal}              & \textbf{98.6}/96.1 & 68.5/\underline{87.2} & 83.7/97.3 & 93.8/90.5 & 59.6/86.5 & \textbf{87.8}/86.6 & 85.3/96.4 & 70.5/94.6 & 89.3/95.5 & 56.1/93.1 & 79.3/92.4 \\
& & CFM~\cite{costanzino2024multimodal}               & 95.6/\underline{96.7} & 62.2/82.9 & 86.1/\underline{98.0} & \underline{96.7}/93.3 & \textbf{84.3}/\textbf{94.2} & 74.6/\underline{93.2} & 72.0/96.2 & 75.8/\underline{97.7} & 95.7/96.8 & 65.2/93.3 & 80.8/93.2 \\
& & MAFR~\cite{ali20252d}              & 94.7/\underline{96.7} & 53.4/84.5 & \underline{92.5}/97.8 & \textbf{96.9}/91.8 & \underline{81.2}/92.2 & 82.2/\textbf{95.2} & 85.3/\underline{97.5} & \underline{77.6}/97.3 & \textbf{97.3}/\underline{96.9} & \textbf{79.4}/\textbf{96.1} & \underline{84.1}/\underline{94.1} \\
\rowcolor{cvprblue!15} 
\cellcolor{white} & \cellcolor{white} & \textit{\textbf{Ours}}     & \underline{97.4}/\textbf{97.3} & \textbf{72.2}/\textbf{92.2} & \textbf{94.9}/\textbf{98.2} & 95.8/93.7 & 74.6/\underline{94.1} & \underline{85.9}/93.0 & \textbf{93.6}/\textbf{98.2} & \textbf{91.2}/\textbf{98.2} & \underline{96.7}/\textbf{97.6} & 68.6/\underline{95.9} & \textbf{87.1}/\textbf{95.8} \\
\bottomrule
\end{tabular}
}
\end{table*}

%===============================EyeCandies Result Updated==============================
\begin{table*}[t!]
\centering
\caption{Few-shot performance comparison on the EyeCandies dataset. The metrics reported are I-AUROC / AUPRO@30\%. The best results are highlighted in \textbf{bold}, and the second-best are \underline{underlined}.}
\label{tab:few_shot_eyecandies}
\resizebox{\textwidth}{!}{
\renewcommand\arraystretch{1.1}
\begin{tabular}{c|c|l ccccccccccc}
\toprule
\rowcolor{gray!20}
Setting & Type & Method & Can. C. & Cho. C. & Cho. P. & Conf. & Gum. B. & Haz. T. & Lic. S. & Lollip. & Marsh. & Pep. C. & Mean \\
\midrule
\multirow{5}{*}{\rotatebox{90}{1-shot}} 
& TF 
  & CIF~\cite{lin2026commonality}               & 38.1/88.4 & \textbf{81.8}/71.4 & 70.2/56.3 & 84.0/86.9 & 59.6/63.9 & 59.7/50.6 & \underline{58.1}/55.7 & \underline{68.6}/\underline{85.8} & 89.1/65.7 & \underline{85.4}/67.1 & \underline{69.5}/69.2 \\
\cmidrule{2-14}
& \multirow{4}{*}{TB} 
  & M3DM~\cite{wang2023multimodal}              & 36.2/86.8 & \underline{66.9}/82.5 & 73.1/\underline{70.6} & 84.8/\underline{94.2} & \underline{71.3}/74.9 & 50.2/54.8 & 57.4/\underline{71.0} & 59.9/84.2 & 60.3/\underline{89.8} & 81.0/\textbf{90.0} & 64.1/\underline{79.9} \\
& & CFM~\cite{costanzino2024multimodal}               & \underline{49.6}/90.1 & 55.8/\underline{83.8} & 61.4/70.1 & \underline{87.0}/90.7 & 65.4/\underline{79.1} & \underline{73.4}/\underline{56.4} & 55.5/61.7 & 52.2/85.5 & 67.5/79.0 & 64.2/78.9 & 63.2/77.5 \\
& & MAFR~\cite{ali20252d}              & \textbf{50.1}/\underline{91.4} & 64.8/82.8 & \underline{76.2}/60.9 & 86.9/\textbf{94.6} & 38.9/78.3 & 70.9/53.9 & 48.6/63.1 & 58.3/83.9 & \underline{92.5}/88.4 & 59.4/80.3 & 64.7/77.8 \\
\rowcolor{cvprblue!15} 
\cellcolor{white} & \cellcolor{white} & \textbf{\textit{Ours}} & 42.1/\textbf{92.6} & 57.9/\textbf{89.3} & \textbf{88.8}/\textbf{79.9} & \textbf{90.6}/90.5 & \textbf{84.1}/\textbf{84.3} & \textbf{74.7}/\textbf{66.5} & \textbf{81.0}/\textbf{82.5} & \textbf{69.9}/\textbf{87.1} & \textbf{94.9}/\textbf{92.7} & \textbf{87.8}/\underline{89.4} & \textbf{77.2}/\textbf{85.5} \\

\midrule
\multirow{5}{*}{\rotatebox{90}{2-shot}} 
& TF 
  & CIF~\cite{lin2026commonality}               & 41.8/87.0 & 79.0/76.9 & 78.6/57.9 & \textbf{94.1}/85.5 & 71.3/61.0 & 70.9/52.4 & 55.7/59.9 & \underline{66.4}/\underline{86.8} & 84.5/71.8 & \textbf{93.6}/74.3 & \underline{73.6}/71.3 \\
\cmidrule{2-14}
& \multirow{4}{*}{TB} 
  & M3DM~\cite{wang2023multimodal}              & 38.9/81.1 & 67.5/84.4 & \underline{81.1}/68.6 & \underline{92.3}/\textbf{97.3} & 61.7/76.1 & 53.6/57.6 & 59.4/\underline{72.8} & 64.0/85.6 & 76.5/89.3 & 86.9/\textbf{90.8} & 68.2/80.4 \\
& & CFM~\cite{costanzino2024multimodal}               & \underline{49.8}/91.3 & \textbf{84.3}/\underline{85.2} & 74.1/\underline{74.3} & 87.7/92.3 & \underline{77.1}/79.6 & 64.2/\textbf{67.7} & \underline{59.8}/68.5 & 54.0/83.9 & 86.6/\underline{91.0} & 63.8/83.4 & 70.1/\underline{81.7} \\
& & MAFR~\cite{ali20252d}              & \textbf{52.0}/\underline{92.0} & 70.2/84.5 & 77.1/69.0 & 87.5/\underline{95.2} & 64.1/\underline{80.8} & \textbf{75.7}/59.1 & 53.3/65.0 & 62.3/84.0 & \underline{91.2}/90.9 & 75.4/86.8 & 70.9/80.7 \\
\rowcolor{cvprblue!15} 
\cellcolor{white} & \cellcolor{white} & \textbf{\textit{Ours}} & 42.9/\textbf{93.3} & \underline{80.5}/\textbf{90.9} & \textbf{90.2}/\textbf{81.0} & 88.0/92.0 & \textbf{84.8}/\textbf{84.8} & \underline{75.0}/\underline{65.8} & \textbf{81.1}/\textbf{80.9} & \textbf{72.2}/\textbf{88.5} & \textbf{98.6}/\textbf{93.5} & \underline{88.6}/\underline{87.7} & \textbf{80.2}/\textbf{85.8} \\

\midrule
\multirow{5}{*}{\rotatebox{90}{4-shot}} 
& TF 
  & CIF~\cite{lin2026commonality}               & \underline{48.2}/87.4 & 81.8/78.1 & 74.6/58.4 & \textbf{97.9}/88.9 & 61.9/68.0 & 69.8/58.1 & 73.3/63.9 & 61.8/88.1 & 94.2/86.0 & 87.4/78.8 & 75.1/75.6 \\
\cmidrule{2-14}
& \multirow{4}{*}{TB} 
  & M3DM~\cite{wang2023multimodal}              & 42.4/82.4 & 74.9/84.9 & 78.7/72.1 & \underline{91.4}/\textbf{96.8} & 70.2/80.2 & 53.4/61.7 & \underline{80.2}/\underline{81.6} & 67.8/\underline{88.6} & 86.6/\textbf{94.6} & \textbf{90.6}/\textbf{92.1} & 73.6/\underline{83.5} \\
& & CFM~\cite{costanzino2024multimodal}               & 47.5/90.9 & \underline{85.3}/85.5 & 76.5/\underline{72.6} & 90.6/94.4 & 76.9/80.0 & 62.6/\underline{68.5} & 65.9/73.5 & \textbf{70.6}/85.7 & 92.2/93.2 & 78.1/87.1 & 74.6/83.1 \\
& & MAFR~\cite{ali20252d}              & \textbf{49.9}/\underline{92.6} & 74.7/\underline{87.3} & \underline{81.1}/70.8 & 87.5/\underline{96.0} & \underline{79.3}/\underline{84.8} & \underline{77.6}/62.4 & 61.3/73.3 & \underline{68.6}/84.9 & \underline{96.0}/93.1 & 80.5/\underline{90.2} & \underline{75.7}/\underline{83.5} \\
\rowcolor{cvprblue!15} 
\cellcolor{white} & \cellcolor{white} & \textbf{\textit{Ours}} & 45.0/\textbf{94.9} & \textbf{91.7}/\textbf{91.3} & \textbf{95.0}/\textbf{81.7} & 88.6/92.6 & \textbf{80.6}/\textbf{86.9} & \textbf{85.4}/\textbf{71.6} & \textbf{85.6}/\textbf{85.8} & 67.7/\textbf{89.0} & \textbf{98.1}/\underline{94.3} & \underline{89.1}/89.1 & \textbf{82.7}/\textbf{87.7} \\
\bottomrule
\end{tabular}
}
\end{table*}

\subsection{Quantitative Comparison}
\label{sec:quantitative_results}

\noindent \textbf{Performance on MVTec 3D-AD.} 
Table \ref{tab:few_shot_combined_updated} shows our framework consistently outperforms all baselines across few-shot settings. In the 1-shot scenario, it achieves 79.6\% I-AUROC and 94.2\% AUPRO@30\%, substantially exceeding the second-best M3DM. Under the 4-shot setting, performance scales to 87.1\% I-AUROC and 95.8\% AUPRO@30\%, significantly surpassing MAFR. Notably, it demonstrates remarkable robustness on geometrically complex categories (\eg, \textit{Bagel}, \textit{Rope}), securing near-perfect localization. These consistent gains highlight the superior cross-modal alignment and generalization enabled by our dual-stream representations.

\noindent \textbf{Performance on EyeCandies.} 
As shown in Table \ref{tab:few_shot_eyecandies}, our approach establishes a new state-of-the-art across all shot settings, validating its strong generalizability. In the 1-shot setting, we achieve 77.2\% I-AUROC and 85.5\% AUPRO@30\%, eclipsing the prior best CIF (69.5\% and 69.2\%, respectively). This superiority persists at 4-shot (82.7\% I-AUROC), demonstrating precise fine-grained anomaly localization under severe data scarcity. Overall, results on both benchmarks confirm that our diffusion-enhanced dual-stream architecture effectively addresses the fundamental bottlenecks of few-shot 3D anomaly detection.

\subsection{Qualitative Analysis}
\label{sec:qualitative}

\begin{figure}[t!]
    \centering
    \includegraphics[width=1.0\linewidth]{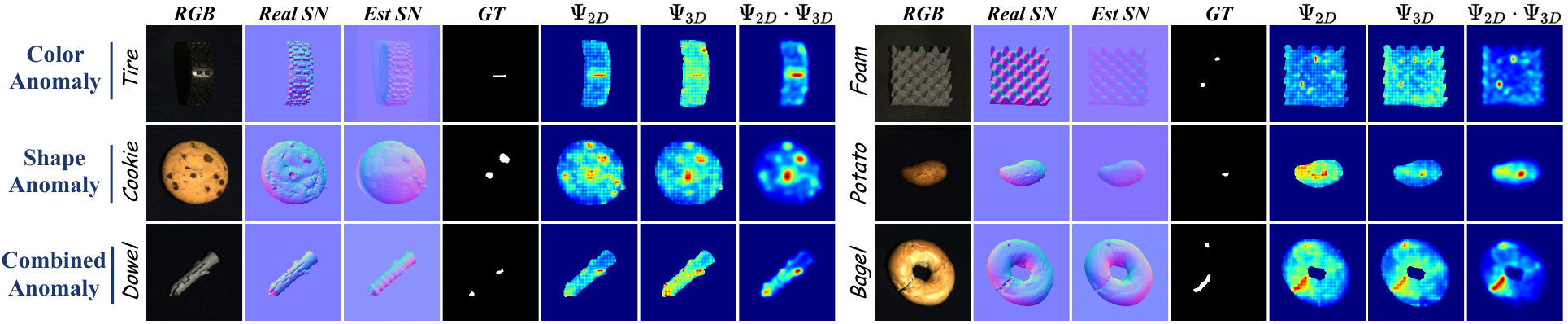}
    \caption{Qualitative evaluation of the complementarity between 2D and 3D modalities. We visualize the predicted anomaly maps across three distinct defect categories: 2D color-only anomalies, 3D shape-only anomalies, and combined anomalies. Synergistically fusing these modality-specific predictions effectively captures all defect types while suppressing background noise.}
    \label{fig:anomaly_types}
\end{figure}

\noindent \textbf{Complementarity of 2D and 3D Modalities.} 
Figure \ref{fig:anomaly_types} visualizes predicted anomaly maps across three distinct defect categories: 2D color-only, 3D shape-only (\eg bumps or dents), and combined anomalies. As illustrated, 2D-focused predictions effectively isolate subtle textural variations, such as surface discolorations or contaminations, which lack geometric deformation. Conversely, 3D-focused predictions precisely highlight structural irregularities that are often imperceptible in the RGB domain. By synergistically fusing these modality-specific outputs, our combined anomaly map robustly localizes complex defects while substantially suppressing modality-specific background noise. This underscores the strong complementarity of our learned 2D and 3D representations.

% ================================== 2D & 3D Modalitied==========================

\begin{figure}[t!]
    \centering
    \includegraphics[width=1.0\linewidth]{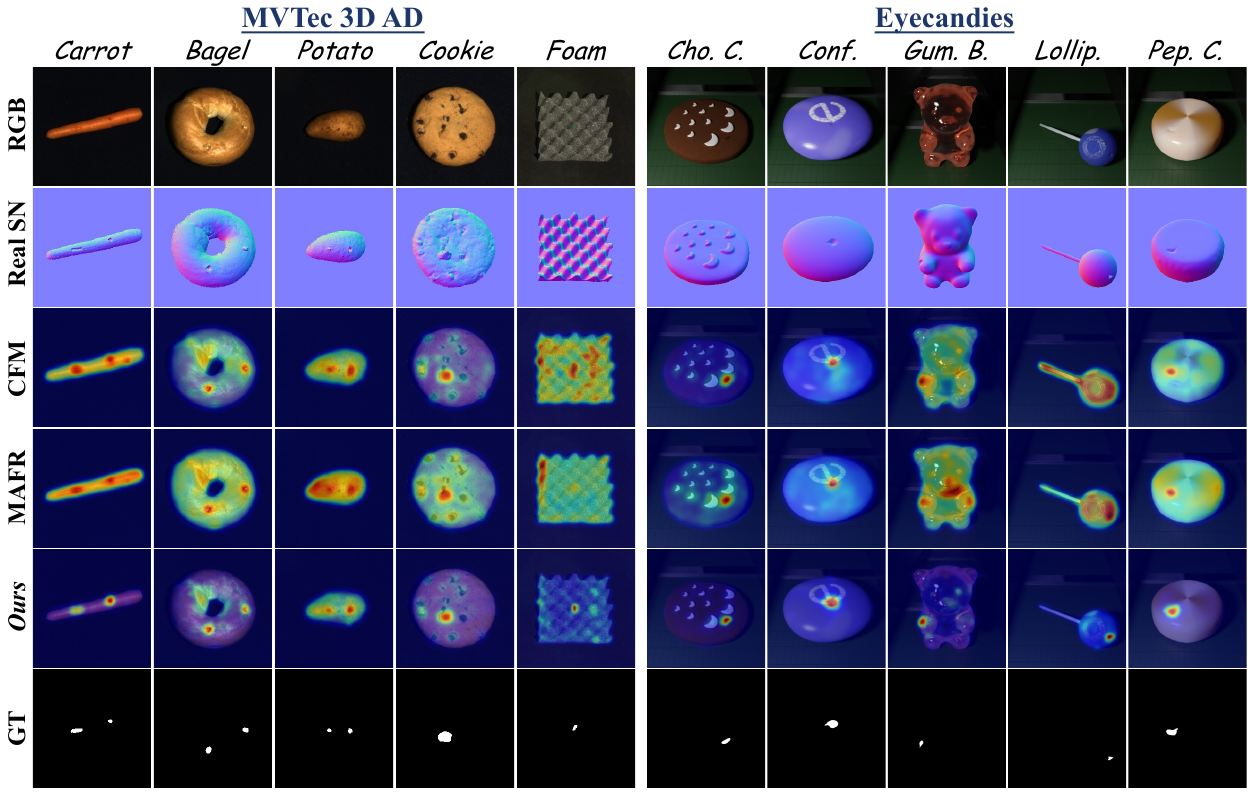}
    \caption{Qualitative comparison of anomaly score overlays between our proposed method and state-of-the-art baselines on the MVTec 3D-AD and EyeCandies datasets. Under challenging few-shot constraints, our approach effectively suppresses false positives in normal regions and delineates subtle defect boundaries with significantly higher precision than existing methods.}
    \label{fig:comparison_overlay}
\end{figure}

\noindent \textbf{Qualitative Comparison with Baselines.} 
To further validate our approach, Fig. \ref{fig:comparison_overlay} visualizes anomaly score overlays for selected categories from the MVTec 3D-AD and EyeCandies datasets against state-of-the-art methods. While existing baselines frequently trigger false alarms in normal regions or struggle to cleanly delineate subtle defect boundaries, our method yields sharp, highly localized anomaly masks that align tightly with the ground truth. Specifically, the diffusion-enhanced structural priors enable our model to maintain high confidence precisely on defective pixels without bleeding into surrounding normal areas. This superior qualitative performance corroborates our quantitative gains, highlighting the robustness and precision of our framework under challenging few-shot constraints.

% =================================== Compare with baseline =======================

% ======================= Ablation Study ===========================
\begin{table}[t!]
\centering
\caption{Ablation study on the core components of the proposed framework under the \textbf{4-shot} setting. The checkmark ($\checkmark$) indicates the inclusion of a specific component. When the proposed Feature Mapper is omitted, a standard MLP is utilized for fusion. The best results are highlighted in \textbf{bold}, and the second-best are \underline{underlined}.}
\label{tab:components_ablation}
% 使用 \tabcolsep 拉开列与列之间（文字与文字之间）的横向空隙
\setlength{\tabcolsep}{1.5mm} 
\resizebox{0.95\columnwidth}{!}{ 
% 使用 \arraystretch 拉开行与行之间的纵向空隙
\renewcommand\arraystretch{1.3} 
\begin{tabular}{ccc | cccccc}
\toprule
\rowcolor{gray!20}
Real Stream & Est Stream & Feature Mapper & I-AUROC & P-AUROC & AUPRO@30\% & AUPRO@10\% & AUPRO@5\% & AUPRO@1\% \\
\midrule
\checkmark & \checkmark & & \underline{0.872} & \underline{0.988} & 0.955 & 0.877 & 0.792 & 0.399 \\
\checkmark & & \checkmark & \textbf{0.873} & \textbf{0.989} & \underline{0.956} & \underline{0.880} & \underline{0.795} & \underline{0.400} \\
& \checkmark & \checkmark & 0.813 & 0.979 & 0.932 & 0.831 & 0.732 & 0.350 \\
\midrule
\rowcolor{cvprblue!15} 
\checkmark & \checkmark & \checkmark & 0.871 & \textbf{0.989} & \textbf{0.958} & \textbf{0.886} & \textbf{0.806} & \textbf{0.410} \\
\bottomrule
\end{tabular}
}
\end{table}

\subsection{Ablation Study}
\label{sec:ablation}

\noindent \textbf{Effectiveness of Core Components.} 
We decouple the real stream, estimated (est) stream, and Feature Mapper under the 4-shot setting (Tab. \ref{tab:components_ablation}). Relying on the est stream yields the weakest I-AUROC (0.813), whereas the real stream alone achieves 0.873. Crucially, integrating both unlocks fine-grained localization, boosting AUPRO@1\% from 0.400 (real-only) to 0.410, confirming that generative priors supplement subtle defect boundaries. Replacing our Feature Mapper with a standard MLP degrades all localization metrics, highlighting the necessity of our hierarchical gating for aligning heterogeneous 2D-3D semantic spaces.

\begin{figure}[t!]
    \centering
    \includegraphics[width=1.0\columnwidth]{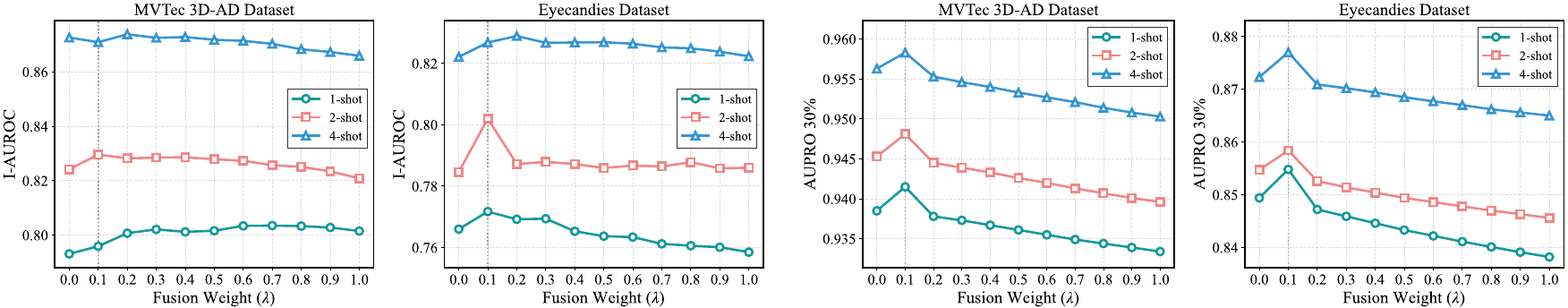}
    \caption{Parameter sensitivity analysis regarding the weights of the estimated stream ($\lambda_1, \lambda_2$) on MVTec 3D-AD and EyeCandies datasets. The plots illustrate the impact on I-AUROC and AUPRO@30\% metrics as the weights vary from 0.0 to 1.0 with a step size of 0.1. From left to right: (a) I-AUROC on MVTec 3D-AD, (b) I-AUROC on EyeCandies, (c) AUPRO@30\% on MVTec 3D-AD, and (d) AUPRO@30\% on EyeCandies.}
    \label{fig:sensitivity_analysis}
\end{figure}

\noindent \textbf{Influence of Estimated Stream Weights.} 
We vary the estimated stream balancing coefficients ($\lambda_1, \lambda_2$) from 0.0 to 1.0 (Fig. \ref{fig:sensitivity_analysis}). Incorporating a moderate generative prior strictly enhances performance: increasing weights from 0.0 to 0.1 improves the 2-shot I-AUROC on EyeCandies (78.46\% to 80.20\%) and the 4-shot AUPRO@30\% on MVTec 3D-AD (95.63\% to 95.83\%). Conversely, over-reliance on estimated features introduces generative noise (\eg at a weight of 1.0, 1-shot I-AUROC on EyeCandies drops to 75.85\%). Thus, an empirical optimal balance is robustly achieved at 0.1 across diverse few-shot settings.

\noindent \textbf{Anomaly Map Fusion Strategy.} 
We evaluate fusion operations for anomaly maps ($\Psi_{\text{2D}}$ and $\Psi_{\text{3D}}$) in Tab. \ref{tab:anomaly_fusion_ablation}. Unimodal detection proves insufficient. Among multi-modal strategies, element-wise multiplication ($\Psi_{\text{2D}} \odot \Psi_{\text{3D}}$) delivers the best localization (0.989 P-AUROC, 0.958 AUPRO@30\%). Multiplication acts as a strict spatial filter, suppressing isolated modality-specific false positives by retaining high scores only when both modalities indicate anomalies. Although addition ($\Psi_{\text{2D}} + \Psi_{\text{3D}}$) yields a marginally higher I-AUROC (0.875), multiplication ensures significantly more robust pixel-level localization under strict false-positive constraints (\eg AUPRO@1\% and 5\%).

% ============================== Sensitivity ===================================

% ================================Different Fusion Method===============================
\begin{table}[t!]
\centering
\caption{Ablation study on the fusion strategy of anomaly maps under the \textbf{4-shot} setting. The best results are highlighted in \textbf{bold}, and the second-best are \underline{underlined}.}
\label{tab:anomaly_fusion_ablation}
\setlength{\tabcolsep}{1.5mm}
\resizebox{0.9\columnwidth}{!}{ % 使用 resizebox 确保表格完美自适应单栏宽度
\renewcommand\arraystretch{1.2}
\begin{tabular}{l | cccccc}
\toprule
\rowcolor{gray!20}
Anomaly Map & I-AUROC & P-AUROC & AUPRO@30\% & AUPRO@10\% & AUPRO@5\% & AUPRO@1\% \\
\midrule
$\Psi_{\text{2D}}$ & 0.807 & 0.985 & 0.948 & 0.864 & 0.773 & 0.372 \\
$\Psi_{\text{3D}}$ & 0.848 & 0.984 & 0.945 & 0.855 & 0.766 & 0.381 \\
$\Psi_{\text{2D}} + \Psi_{\text{3D}}$ & \textbf{0.875} & \underline{0.986} & \underline{0.952} & \underline{0.874} & \underline{0.791} & \underline{0.401} \\
$\max(\Psi_{\text{2D}}, \Psi_{\text{3D}})$ & 0.861 & 0.985 & 0.950 & 0.869 & 0.785 & 0.394 \\
\rowcolor{cvprblue!15} 
$\Psi_{\text{2D}} \odot \Psi_{\text{3D}}$ & \underline{0.871} & \textbf{0.989} & \textbf{0.958} & \textbf{0.886} & \textbf{0.806} & \textbf{0.410} \\
\bottomrule
\end{tabular}
}
\end{table}

\section{Conclusion}
We propose CMDS-AD, a Cross-Modal Dual-Stream framework for few-shot anomaly detection. Repurposing a diffusion estimator as a non-linear low-pass filter establishes a low-frequency auxiliary stream that enhances the real stream, isolating micro-defects from macroscopic structures. A Coordinate-Aware Hierarchical Feature Mapper adaptively bridges cross-modal semantic gaps, and a multiplicative scoring mechanism ($\Psi_{\text{2D}} \odot \Psi_{\text{3D}}$) suppresses modality-specific noise. Experiments on MVTec 3D-AD and EyeCandies show CMDS-AD overcomes data scarcity, achieving state-of-the-art 1- to 4-shot anomaly localization.

% \clearpage  % TODO FINAL: This \clearpage needs to be removed from both review and camera-ready versions.

\section*{Acknowledgements}

This research work was financially supported in part by the Guangdong Major Project of Basic Research under Grant 2023B0303000009, in part by the NSFC Youth Fund Project under Grant 62403326, in part by the Shenzhen Fundamental Research Fund under Grant JCYJ20230808105212023, in part by the Research Team Cultivation Program of ShenZhen University under Grant 2023JCT004, and in part by the Shenzhen University 2035 Program for Excellent Research under Grant 00000224.

\bibliographystyle{splncs04}
\bibliography{main}

% ==========================================
% Appendix / Supplementary Material
% ==========================================
\clearpage % 强制换页，让附录从新的一页开始（强烈建议）
\appendix  % 核心命令：将章节编号从 1, 2, 3... 切换为 A, B, C...
\begin{center}
    \Large \bf Supplementary Material for CMDS-AD: Cross-Modal Dual-Stream Decoupling for Few-Shot Anomaly Detection
\end{center}

\renewcommand{\theHsection}{appendix.\Alph{section}}
\renewcommand{\theHsubsection}{appendix.\Alph{section}.\arabic{subsection}}

\section{Extended Analysis on Strict FPR Thresholds}

\subsection{Evaluation under Strict Few-Shot Settings}

In industrial anomaly detection, achieving precise defect localization with a low False Positive Rate (FPR) is crucial. However, current few-shot multi-modal anomaly detection literature predominantly reports the Area Under the Per-Region-Overlap curve up to a 30\% FPR (AUPRO@30\%). While this metric provides a general sense of localization capability, it may not fully reflect a model's reliability in real-world deployments. Specifically, in high-throughput manufacturing environments, allowing up to a 30\% false positive rate would result in an unacceptable volume of normal items being flagged for review, incurring significant manual re-inspection costs and disrupting production efficiency. 

To better align with practical requirements, it is necessary to evaluate models under stricter FPR thresholds. Evaluating at these constrained limits effectively isolates a model's ability to suppress background noise and handle ambiguous structural variations without triggering false alarms. In Table \ref{tab:strict_fpr}, we present the class-wise performance of our CMDS-AD framework on the MVTec 3D-AD dataset across 1-shot, 2-shot, and 4-shot settings, detailing AUPRO at 10\%, 5\%, and 1\%. By providing these detailed metrics, we aim to offer a more practical reference baseline for future research focused on robust, low-FPR defect localization.

\subsection{PRO Curve Dynamics: The Superiority of Multiplicative Fusion}

\begin{table*}[t!]
\centering
\caption{Comprehensive performance of CMDS-AD under strict FPR thresholds on MVTec 3D-AD. We report AUPRO evaluated at integration limits of 10\%, 5\%, and 1\% to demonstrate robustness in strict industrial scenarios. For each threshold across different shots, the best results are highlighted in \textbf{bold}, and the second-best are \underline{underlined}.}
\label{tab:strict_fpr}
\resizebox{\textwidth}{!}{
\renewcommand\arraystretch{1.1}
\begin{tabular}{l|ccc|ccc|ccc}
\toprule
\multirow{2}{*}{Class} & \multicolumn{3}{c|}{1-shot} & \multicolumn{3}{c|}{2-shot} & \multicolumn{3}{c}{4-shot} \\
\cmidrule(lr){2-4} \cmidrule(lr){5-7} \cmidrule(lr){8-10}
& AUPRO@10\% & AUPRO@5\% & AUPRO@1\% & AUPRO@10\% & AUPRO@5\% & AUPRO@1\% & AUPRO@10\% & AUPRO@5\% & AUPRO@1\% \\
\midrule
Bagel       & 0.915 & 0.842 & 0.427 & \underline{0.920} & \underline{0.851} & \underline{0.436} & \textbf{0.922} & \textbf{0.855} & \textbf{0.442} \\
Cable Gland & 0.679 & 0.507 & 0.179 & \underline{0.702} & \underline{0.535} & \underline{0.192} & \textbf{0.776} & \textbf{0.636} & \textbf{0.268} \\
Carrot      & 0.942 & 0.886 & 0.472 & \underline{0.944} & \underline{0.890} & \underline{0.478} & \textbf{0.947} & \textbf{0.894} & \textbf{0.485} \\
Cookie      & 0.869 & 0.801 & 0.416 & \underline{0.873} & \underline{0.806} & \underline{0.419} & \textbf{0.877} & \textbf{0.810} & \textbf{0.429} \\
Dowel       & 0.657 & 0.461 & 0.150 & \underline{0.681} & \underline{0.497} & \underline{0.166} & \textbf{0.824} & \textbf{0.705} & \textbf{0.312} \\
Foam        & 0.734 & 0.624 & 0.285 & \underline{0.802} & \underline{0.704} & \underline{0.343} & \textbf{0.815} & \textbf{0.722} & \textbf{0.359} \\
Peach       & \underline{0.938} & \underline{0.876} & \underline{0.455} & 0.935 & 0.871 & 0.444 & \textbf{0.944} & \textbf{0.889} & \textbf{0.477} \\
Potato      & 0.939 & 0.879 & 0.451 & \underline{0.944} & \underline{0.888} & \underline{0.475} & \textbf{0.946} & \textbf{0.893} & \textbf{0.482} \\
Rope        & \underline{0.929} & \underline{0.868} & 0.455 & \textbf{0.931} & \textbf{0.872} & \textbf{0.459} & \underline{0.929} & 0.867 & \underline{0.456} \\
Tire        & 0.818 & 0.692 & 0.301 & \underline{0.848} & \underline{0.733} & \underline{0.330} & \textbf{0.878} & \textbf{0.784} & \textbf{0.384} \\
\midrule
\textbf{MEAN} & 0.842 & 0.743 & 0.359 & \underline{0.858} & \underline{0.765} & \underline{0.374} & \textbf{0.886} & \textbf{0.805} & \textbf{0.409} \\
\bottomrule
\end{tabular}
}
\end{table*}

\begin{figure}[t!]
    \centering
    % Please uncomment and adjust the path to your actual figure
    \includegraphics[width=1.0\linewidth]{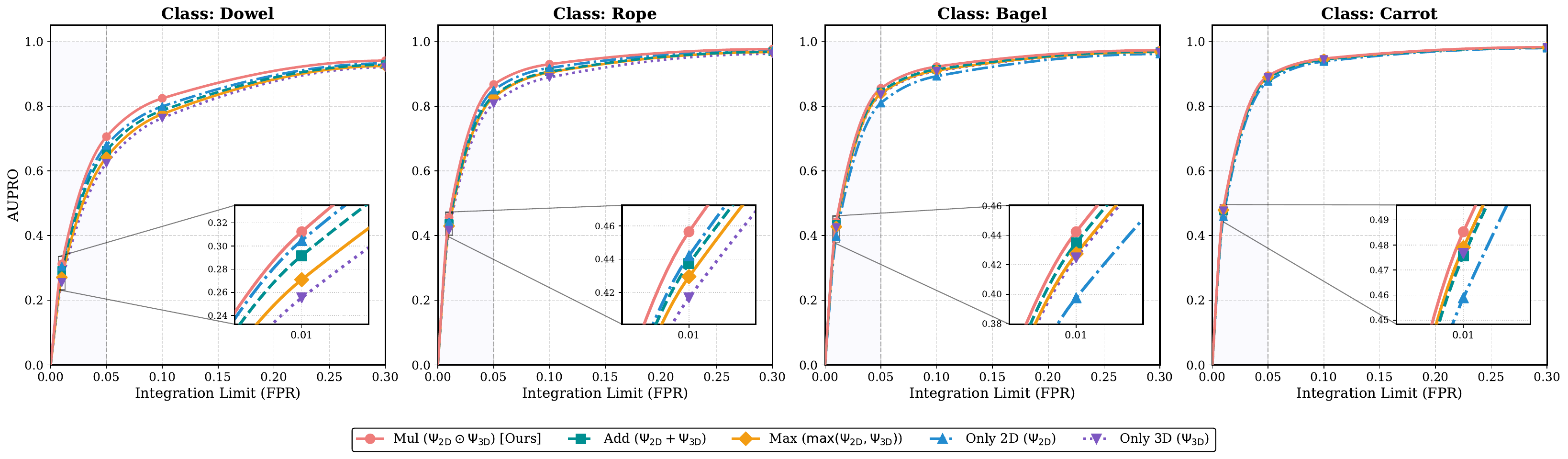}
    \caption{PRO curve dynamics across different integration limits (FPR). We select four representative classes (\textit{Dowel}, \textit{Rope}, \textit{Bagel}, \textit{Carrot}) to illustrate the varying degradation behaviors of different fusion strategies. The inset panels provide a 10$\times$ magnified view specifically anchored at the strict 1\% FPR threshold. Multiplicative fusion ($\Psi_{\text{2D}} \odot \Psi_{\text{3D}}$) consistently maintains top-tier stability, effectively resisting the unilateral noise that severely degrades single-stream, Additive, and Max fusion strategies.}
    \label{fig:pro_dynamics}
\end{figure}

To further elucidate the mechanisms behind our model's robustness under strict industrial constraints, we analyze the continuous PRO curve dynamics across different fusion strategies in Figure \ref{fig:pro_dynamics}. We select four geometrically and texturally challenging classes (\textit{Dowel}, \textit{Rope}, \textit{Bagel}, and \textit{Carrot}) as representative cases, as they encompass a diverse spectrum of complex local curvatures, highly variable surface reflectivity, and non-deterministic structural boundaries.

As observed in the macroscopic view, at looser constraints approaching 30\% FPR, all tested fusion strategies, including additive ($\Psi_{\text{2D}} + \Psi_{\text{3D}}$), maximum ($\max(\Psi_{\text{2D}}, \Psi_{\text{3D}})$), and multiplicative ($\Psi_{\text{2D}} \odot \Psi_{\text{3D}}$), yield ostensibly comparable and near-saturated performance. However, as the FPR threshold is restricted to the extreme low end (the highlighted $\le$ 5\% zone), a distinct bifurcation occurs. Modality-specific noise, such as specular highlights in 2D (e.g., on the \textit{Bagel} surface) or inherent sensor artifacts in 3D (e.g., on the thin structure of \textit{Dowel}), causes the single-modality branches to experience a steep, cliff-like drop in precision. Without a mechanism to critically cross-reference conflicting signals, these traditional strategies are easily overwhelmed by localized but entirely normal variations.

Crucially, as highlighted in the inset zoom panels at exactly 1\% FPR, both Additive and Max fusion strategies fail to suppress this degradation, as they are highly susceptible to confident false positives originating from any single noisy modality. In stark contrast, our Multiplicative fusion acts as a stringent logical spatial \textit{AND} filter. A localized anomaly is only strongly activated if corroborated by both the 2D texture variation and the 3D structural deviation simultaneously. This explicit cross-modal verification effectively neutralizes unilateral noise, allowing CMDS-AD to maintain a resilient and dominant PRO curve trajectory even at ultra-low FPRs, ensuring that the model exclusively flags regions with genuine multimodal defect evidence rather than transient background artifacts.

\subsection{Why Single-Stream Architecture Collapses at Low FPRs}

The divergence in performance under strict FPRs can be deeply understood through the lens of our dual-stream architecture (Real Stream vs. Estimation Stream). As demonstrated in our ablation studies, relying exclusively on the Real Stream (which inherently contains both high and low-frequency spatial variations) allows the model to perform adequately at AUPRO@30\% (0.9563). However, under the extreme pressure of AUPRO@1\%, its performance collapses to 0.4002. 

The underlying mechanism for this failure is the model's over-sensitivity to normal, localized high-frequency texture variations, which it misclassifies as micro-defects when forced to minimize false positives. By introducing the Estimation Stream—where the diffusion model acts as an advanced low-pass filter—we extract stabilized low-frequency priors. These priors serve as structural anchors. The explicit decoupling of high-frequency details (Real) and macro-structural semantics (Estimation) is the decisive factor that inhibits false positives and overcomes the data-scarcity bottleneck inherent in few-shot settings.

\vspace{-2pt}

\section{Comprehensive Class-wise Ablation and P-AUROC Results}

\begin{table*}[t!]
\centering
\caption{Comprehensive class-wise ablation of Core Components (4-shot, MVTec 3D-AD). Corresponding to Table 3 in the main text. The best results are highlighted in \textbf{bold}, and the second-best are \underline{underlined}. Note the stable and saturated P-AUROC values across all configurations.}
\label{tab:ablation_components_full}
\resizebox{\textwidth}{!}{
\begin{tabular}{l|ccc|ccc|ccc|ccc}
\toprule
\multirow{2}{*}{Class} & \multicolumn{3}{c|}{w/o Est Stream} & \multicolumn{3}{c|}{w/o Real Stream} & \multicolumn{3}{c|}{w/o Feature Mapper} & \multicolumn{3}{c}{\textbf{Ours (Dual-Stream)}} \\
\cmidrule(lr){2-4} \cmidrule(lr){5-7} \cmidrule(lr){8-10} \cmidrule(lr){11-13}
& PRO@30\% & PRO@1\% & P-AUC & PRO@30\% & PRO@1\% & P-AUC & PRO@30\% & PRO@1\% & P-AUC & PRO@30\% & PRO@1\% & P-AUC \\
\midrule
Bagel       & \underline{0.968} & 0.404 & \textbf{0.994} & 0.964 & \underline{0.421} & 0.990 & \underline{0.968} & 0.409 & \underline{0.993} & \textbf{0.972} & \textbf{0.442} & \textbf{0.994} \\
Cable Gland & \underline{0.916} & 0.254 & \underline{0.972} & 0.907 & \underline{0.259} & 0.969 & 0.907 & 0.251 & 0.969 & \textbf{0.922} & \textbf{0.268} & \textbf{0.974} \\
Carrot      & \textbf{0.982} & \textbf{0.485} & \textbf{0.998} & \underline{0.980} & \underline{0.470} & \underline{0.997} & \textbf{0.982} & \textbf{0.485} & \textbf{0.998} & \textbf{0.982} & \textbf{0.485} & \textbf{0.998} \\
Cookie      & \textbf{0.939} & \textbf{0.430} & \textbf{0.971} & 0.876 & 0.290 & 0.948 & 0.930 & 0.417 & 0.968 & \underline{0.936} & \underline{0.429} & \underline{0.970} \\
Dowel       & \underline{0.937} & \underline{0.297} & \underline{0.984} & 0.910 & 0.226 & 0.976 & 0.934 & 0.296 & \underline{0.984} & \textbf{0.940} & \textbf{0.312} & \textbf{0.985} \\
Foam        & \underline{0.924} & 0.346 & \underline{0.981} & 0.829 & 0.289 & 0.943 & \textbf{0.929} & \underline{0.353} & \textbf{0.982} & \textbf{0.929} & \textbf{0.359} & \textbf{0.982} \\
Peach       & \textbf{0.981} & \textbf{0.477} & \textbf{0.998} & \underline{0.979} & 0.467 & \underline{0.997} & \textbf{0.981} & \underline{0.473} & \textbf{0.998} & \textbf{0.981} & \textbf{0.477} & \textbf{0.998} \\
Potato      & \textbf{0.982} & 0.479 & \textbf{0.998} & \underline{0.971} & 0.370 & \underline{0.992} & \textbf{0.982} & \underline{0.481} & \textbf{0.998} & \textbf{0.982} & \textbf{0.482} & \textbf{0.998} \\
Rope        & \textbf{0.976} & \textbf{0.458} & \textbf{0.997} & 0.970 & 0.440 & \underline{0.996} & \underline{0.975} & 0.455 & \textbf{0.997} & \textbf{0.976} & \underline{0.456} & \textbf{0.997} \\
Tire        & \underline{0.955} & \underline{0.367} & \underline{0.990} & 0.925 & 0.260 & 0.979 & \underline{0.955} & 0.366 & 0.989 & \textbf{0.958} & \textbf{0.384} & \textbf{0.991} \\
\midrule
\textbf{MEAN} & \underline{0.956} & \underline{0.400} & \underline{0.988} & 0.931 & 0.349 & 0.979 & 0.954 & 0.399 & \underline{0.988} & \textbf{0.958} & \textbf{0.409} & \textbf{0.989} \\
\bottomrule
\end{tabular}
}
\end{table*}

Due to space constraints in the main manuscript, Tables 3 and 4 reported the mean metrics across all classes. To provide a completely transparent view of our model's behavior, Tables \ref{tab:ablation_components_full} and \ref{tab:ablation_fusion_full} present the detailed class-wise breakdown for the core component and fusion strategy ablations, respectively (evaluated at 4-shot on MVTec 3D-AD). 

Furthermore, these tables incorporate the pixel-level anomaly detection metric (P-AUROC). As observed in recent state-of-the-art literature, P-AUROC has largely saturated (consistently hovering between 98\% and 99\% across most configurations). Thus, while it is no longer a discriminative metric for assessing fine-grained localization superiority, we report it fully in this appendix to facilitate alignment with earlier literature.

More importantly, the granular data in Table \ref{tab:ablation_components_full} explicitly reveals the critical role of the dual-branch decoupling mechanism. For instance, when omitting the Estimation Stream (\textit{w/o Est Stream}), the model still achieves a near-perfect P-AUROC (0.988) and a competitive AUPRO@30\% (0.956), creating an illusion of high performance. However, its AUPRO@1\% plummets to 0.400 across all classes. This validates our core hypothesis: without the low-frequency structural anchor provided by the diffusion model, the network inevitably overfits to normal high-frequency texture variations during few-shot learning, leading to a catastrophic collapse in precise localization when false positives are strictly penalized. The detailed data unequivocally demonstrates that our full CMDS-AD framework maintains exceptional stability across all diverse and geometrically challenging object categories.

\begin{table*}[t!]
\centering
\caption{Comprehensive class-wise ablation of Fusion Strategies (4-shot, MVTec 3D-AD). Corresponding to Table 4 in the main text. The best results are highlighted in \textbf{bold}, and the second-best are \underline{underlined}.}
\label{tab:ablation_fusion_full}
\resizebox{\textwidth}{!}{
\begin{tabular}{l|ccc|ccc|ccc|ccc|ccc}
\toprule
\multirow{2}{*}{Class} & \multicolumn{3}{c|}{Only 2D} & \multicolumn{3}{c|}{Only 3D} & \multicolumn{3}{c|}{Max} & \multicolumn{3}{c|}{Add} & \multicolumn{3}{c}{\textbf{Mul (Ours)}} \\
\cmidrule(lr){2-4} \cmidrule(lr){5-7} \cmidrule(lr){8-10} \cmidrule(lr){11-13} \cmidrule(lr){14-16}
& PRO@30 & PRO@1 & P-AUC & PRO@30 & PRO@1 & P-AUC & PRO@30 & PRO@1 & P-AUC & PRO@30 & PRO@1 & P-AUC & PRO@30 & PRO@1 & P-AUC \\
\midrule
Bagel       & 0.961 & 0.397 & 0.991 & 0.966 & 0.424 & 0.990 & 0.967 & 0.427 & 0.991 & \underline{0.968} & \underline{0.434} & \underline{0.992} & \textbf{0.972} & \textbf{0.442} & \textbf{0.994} \\
Cable Gland & \textbf{0.931} & \textbf{0.288} & \textbf{0.977} & 0.885 & 0.216 & 0.963 & 0.913 & 0.245 & 0.972 & 0.914 & 0.258 & 0.972 & \underline{0.922} & \underline{0.268} & \underline{0.974} \\
Carrot      & 0.979 & 0.458 & 0.996 & \underline{0.981} & 0.476 & \textbf{0.998} & \underline{0.981} & \underline{0.479} & \textbf{0.998} & \underline{0.981} & 0.475 & \underline{0.997} & \textbf{0.982} & \textbf{0.485} & \textbf{0.998} \\
Cookie      & 0.897 & 0.327 & 0.951 & 0.910 & 0.419 & 0.953 & 0.907 & 0.413 & 0.951 & \underline{0.911} & \underline{0.428} & \underline{0.954} & \textbf{0.936} & \textbf{0.429} & \textbf{0.970} \\
Dowel       & \underline{0.932} & \underline{0.304} & \underline{0.982} & 0.920 & 0.255 & 0.979 & 0.924 & 0.271 & 0.980 & 0.928 & 0.291 & 0.981 & \textbf{0.940} & \textbf{0.312} & \textbf{0.985} \\
Foam        & 0.912 & 0.326 & 0.975 & 0.909 & 0.289 & 0.977 & \underline{0.920} & 0.348 & 0.978 & \textbf{0.929} & \underline{0.355} & \underline{0.981} & \textbf{0.929} & \textbf{0.359} & \textbf{0.982} \\
Peach       & 0.975 & 0.423 & \underline{0.995} & \textbf{0.981} & \underline{0.476} & \textbf{0.998} & \underline{0.980} & 0.471 & \textbf{0.998} & \textbf{0.981} & 0.474 & \textbf{0.998} & \textbf{0.981} & \textbf{0.477} & \textbf{0.998} \\
Potato      & 0.974 & 0.410 & 0.994 & \textbf{0.982} & \textbf{0.484} & \textbf{0.998} & \textbf{0.982} & 0.481 & \textbf{0.998} & \underline{0.981} & 0.476 & \underline{0.997} & \textbf{0.982} & \underline{0.482} & \textbf{0.998} \\
Rope        & \underline{0.972} & \underline{0.442} & \textbf{0.997} & 0.961 & 0.416 & 0.995 & 0.967 & 0.429 & \underline{0.996} & 0.968 & 0.437 & \underline{0.996} & \textbf{0.976} & \textbf{0.456} & \textbf{0.997} \\
Tire        & 0.948 & 0.341 & 0.987 & 0.945 & 0.354 & 0.987 & 0.951 & 0.372 & 0.988 & \underline{0.955} & \underline{0.380} & \underline{0.989} & \textbf{0.958} & \textbf{0.384} & \textbf{0.991} \\
\midrule
\textbf{MEAN} & 0.948 & 0.372 & 0.985 & 0.944 & 0.381 & 0.984 & 0.949 & 0.394 & 0.985 & \underline{0.951} & \underline{0.401} & \underline{0.986} & \textbf{0.958} & \textbf{0.409} & \textbf{0.989} \\
\bottomrule
\end{tabular}
}
\end{table*}

\section{Mechanistic Validation and Robustness Analysis}

\subsection{Spectral Evidence for Frequency-Decoupled Priors}

\begin{figure}[t!]
    \centering
    \includegraphics[width=0.85\linewidth]{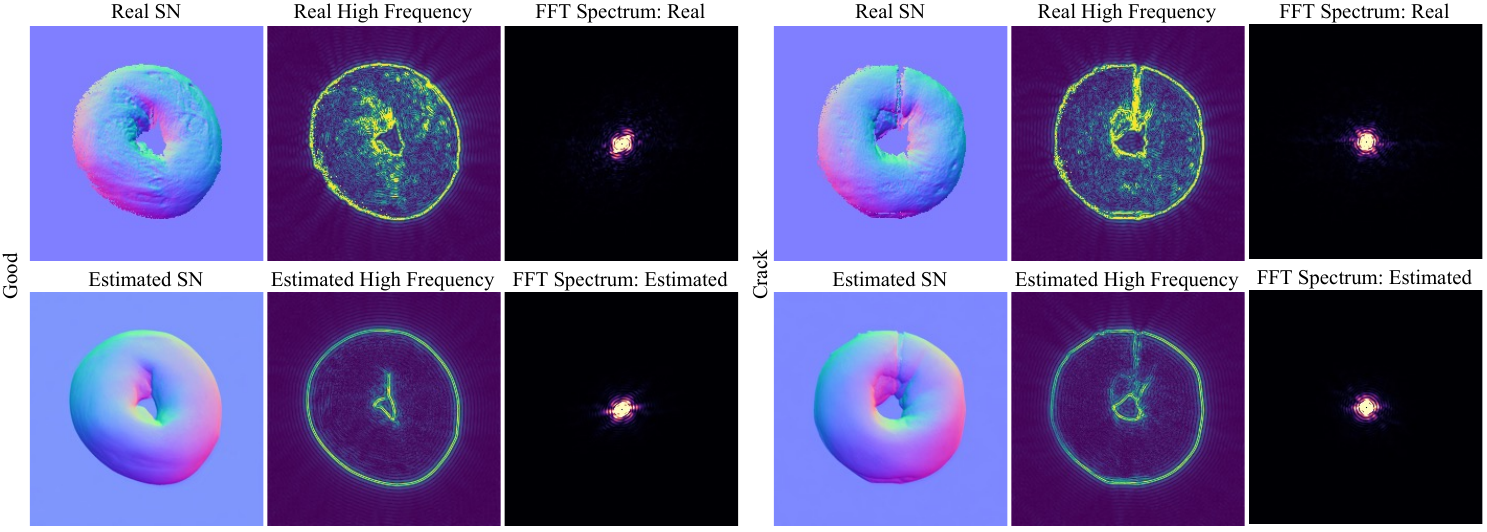}
    \caption{FFT-based frequency analysis of real and estimated normals. Estimated normals suppress high-frequency responses while preserving dominant structural components.}
    \label{fig:rebuttal_frequency_appendix}
\end{figure}

To directly substantiate the frequency-decoupling interpretation, Figure \ref{fig:rebuttal_frequency_appendix} visualizes the FFT spectra of real and estimated normals, while Table \ref{tab:spectral_metrics_appendix} reports the corresponding normalized spectral energy statistics. The estimated normals clearly suppress high-frequency responses and retain coarse structural components. Quantitatively, the High/Low energy ratio drops from 0.1251 to 0.0221 on normal samples and from 0.1293 to 0.0239 on anomalous samples. Together, these results support our view that the diffusion-based normal estimator acts as a non-linear low-pass prior: it preserves stable global structure while suppressing modality-specific high-frequency noise that would otherwise trigger false positives.

\begin{table}[t!]
    \centering
    \caption{Spectral energy distribution of real and estimated normals. Estimated normals exhibit substantially lower high-frequency energy and a markedly smaller High/Low ratio.}
    \label{tab:spectral_metrics_appendix}
    \setlength{\tabcolsep}{3.5pt}
    \resizebox{0.85\linewidth}{!}{
        \begin{tabular}{lcccccc}
            \toprule
            \multirow{2}{*}{Method} & \multicolumn{3}{c}{Normal Samples} & \multicolumn{3}{c}{Anomaly Samples} \\
            \cmidrule(lr){2-4} \cmidrule(lr){5-7}
             & Low E. & High E. & H/L Ratio & Low E. & High E. & H/L Ratio \\
            \midrule
            Real Normal & 0.8888 & 0.1112 & 0.1251 & 0.8855 & 0.1145 & 0.1293 \\
            Estimated Normal & 0.9784 & 0.0216 & 0.0221 & 0.9766 & 0.0234 & 0.0239 \\
            \bottomrule
        \end{tabular}
    }
\end{table}

\subsection{Isolating Architectural Gains Beyond Shared Diffusion Priors}

The component ablations above validate the role of each module within CMDS-AD, but they do not fully separate architectural gains from the strength of the shared diffusion priors. To disentangle these factors, Table \ref{tab:appendix_prior_control} compares our method against an ``MLP'' baseline that uses the same Stable Diffusion v2.1 augmentation and Marigold-estimated normals, but replaces the dual-stream design with a shallow multimodal predictor. Even under identical priors, CMDS-AD still delivers consistent gains, especially at strict operating points such as PRO@1\% (+1.1 points over MLP). This shows that the benefit is not merely inherited from better priors; it is amplified by the proposed dual-stream architecture and its cross-modal interaction design.

\begin{table}[t!]
\centering
\caption{Ablation isolating the architectural contribution under identical diffusion priors on MVTec 3D-AD (4-shot). ``MLP'' uses the same Stable Diffusion v2.1 augmentation and Marigold normal estimation as our method.}
\label{tab:appendix_prior_control}
\resizebox{0.8\linewidth}{!}{
\begin{tabular}{lcccccc}
\toprule
Method & I-AUROC & P-AUROC & PRO@30\% & PRO@10\% & PRO@5\% & PRO@1\% \\
\midrule
CFM & 80.8 & 98.5 & 94.2 & 84.3 & 73.9 & 35.2 \\
MAFR & 84.1 & 98.4 & 94.6 & 85.6 & 76.5 & 37.7 \\
\midrule
MLP & \textbf{87.2} & \underline{98.8} & \underline{95.5} & \underline{87.7} & \underline{79.2} & \underline{39.9} \\
\textbf{Ours} & \underline{87.1} & \textbf{98.9} & \textbf{95.8} & \textbf{88.6} & \textbf{80.6} & \textbf{41.0} \\
\bottomrule
\end{tabular}
}
\end{table}

\subsection{Sensitivity to Estimated-Stream Balancing Weights}

To complement the sensitivity curves in the main text, Table \ref{tab:appendix_sensitivity} summarizes the mean performance when the two estimated-stream balancing coefficients are tied as $\lambda_1 = \lambda_2 = w$. Across $w \in [0.1, 0.9]$, P-AUROC remains nearly constant, while PRO@1\% varies within only about 1.6 points on MVTec 3D-AD and 1.4 points on EyeCandies. This indicates that the method does not require delicate tuning, and the default choice $w=0.1$ used in the main manuscript remains a robust operating point across both benchmarks.

\begin{table*}[t!]
\centering
\caption{Sensitivity to the shared estimated-stream balancing weight $w$, where $\lambda_1 = \lambda_2 = w$. Best and second-best are highlighted in \textbf{bold} and \underline{underlined}, respectively.}
\label{tab:appendix_sensitivity}
\setlength{\tabcolsep}{4.5pt}
\resizebox{0.98\textwidth}{!}{
\begin{tabular}{ccccccccc}
\toprule
\multirow{2}{*}{$w$} & \multicolumn{4}{c}{MVTec 3D-AD} & \multicolumn{4}{c}{EyeCandies} \\
\cmidrule(lr){2-5} \cmidrule(lr){6-9}
 & I-AUROC & P-AUROC & PRO@30\% & PRO@1\% & I-AUROC & P-AUROC & PRO@30\% & PRO@1\% \\
\midrule
$0.1$ & 87.1 & \textbf{98.9} & \textbf{95.8} & \textbf{41.0} & \textbf{82.7} & \textbf{97.4} & \textbf{87.7} & \textbf{30.4} \\
$0.3$ & \textbf{87.3} & \underline{98.8} & \underline{95.5} & \underline{40.1} & \textbf{82.7} & \underline{97.3} & \underline{87.0} & \underline{29.4} \\
$0.5$ & \underline{87.2} & 98.7 & 95.3 & 39.9 & \textbf{82.7} & 97.2 & 86.9 & 29.3 \\
$0.7$ & 87.0 & 98.7 & 95.2 & 39.6 & \underline{82.5} & 97.2 & 86.7 & 29.2 \\
$0.9$ & 86.7 & 98.6 & 95.1 & 39.4 & 82.4 & 97.2 & 86.6 & 29.0 \\
\bottomrule
\end{tabular}
}
\end{table*}

\subsection{Cross-Shot Stability and Synthetic-Count Sensitivity}

Table \ref{tab:appendix_stability} extends the strict-FPR analysis by aggregating baseline comparisons across 1-shot, 2-shot, and 4-shot settings, together with 5-seed statistics and an ablation on the number of synthetic samples $N$ generated per real image. CMDS-AD consistently improves over CFM and MAFR, with the largest margin appearing at the strictest PRO@1\% criterion. Moreover, the 4-shot 5-seed results exhibit low variance ($41.3_{\pm0.8}$ at PRO@1\%), and varying $N$ from 1 to 3 only causes marginal shifts. These observations indicate that the reported gains are stable rather than driven by a favorable seed or a highly sensitive augmentation count.

\begin{table*}[t!]
\centering
\caption{Cross-shot strict-FPR comparison on MVTec 3D-AD, including 4-shot stability over 5 random seeds and an ablation on the number of synthetic samples $N$ generated per real image.}
\label{tab:appendix_stability}
\setlength{\tabcolsep}{3.2pt}
\resizebox{0.7\textwidth}{!}{
\begin{tabular}{clcccccc}
\toprule
Shot & Method & I-AUC & P-AUC & 30\% & 10\% & 5\% & 1\% \\
\midrule
\multirow{3}{*}{1}
& CFM  & 69.8 & 97.5 & 91.3 & 76.8 & 63.4 & 26.2 \\
& MAFR & 72.4 & 97.7 & 92.2 & 79.4 & 67.6 & 30.0 \\
& \textbf{Ours} & \textbf{79.6} & \textbf{98.3} & \textbf{94.2} & \textbf{84.3} & \textbf{74.4} & \textbf{36.0} \\
\midrule
\multirow{3}{*}{2}
& CFM  & 73.0 & 98.0 & 92.6 & 80.3 & 68.5 & 30.6 \\
& MAFR & 76.6 & 98.0 & 93.2 & 81.8 & 71.6 & 33.6 \\
& \textbf{Ours} & \textbf{83.0} & \textbf{98.5} & \textbf{94.8} & \textbf{85.9} & \textbf{76.5} & \textbf{37.5} \\
\midrule
\multirow{6}{*}{4}
& CFM  & 80.8 & 98.5 & 94.2 & 84.3 & 73.9 & 35.2 \\
& MAFR & 84.1 & 98.4 & 94.6 & 85.6 & 76.5 & 37.7 \\
& \textbf{Ours} & \textbf{87.1} & \textbf{98.9} & \textbf{95.8} & \textbf{88.6} & \textbf{80.6} & \textbf{41.0} \\
\cmidrule(lr){2-8}
& 5 seeds & $86.8_{\pm 1.2}$ & $98.9_{\pm 0.1}$ & $95.9_{\pm 0.3}$ & $88.8_{\pm 0.7}$ & $80.9_{\pm 0.9}$ & $41.3_{\pm 0.8}$ \\
& $N=1$ & 86.5 & 98.9 & 95.8 & 88.5 & 80.3 & 40.4 \\
& $N=3$ & 86.2 & 98.9 & 95.9 & 88.7 & 80.5 & 40.6 \\
\bottomrule
\end{tabular}
}
\end{table*}

\section{Extended Qualitative Results and Failure Cases}

\subsection{Additional Qualitative Results}

\begin{figure}[htbp]
    \centering
    \includegraphics[width=0.91\columnwidth]{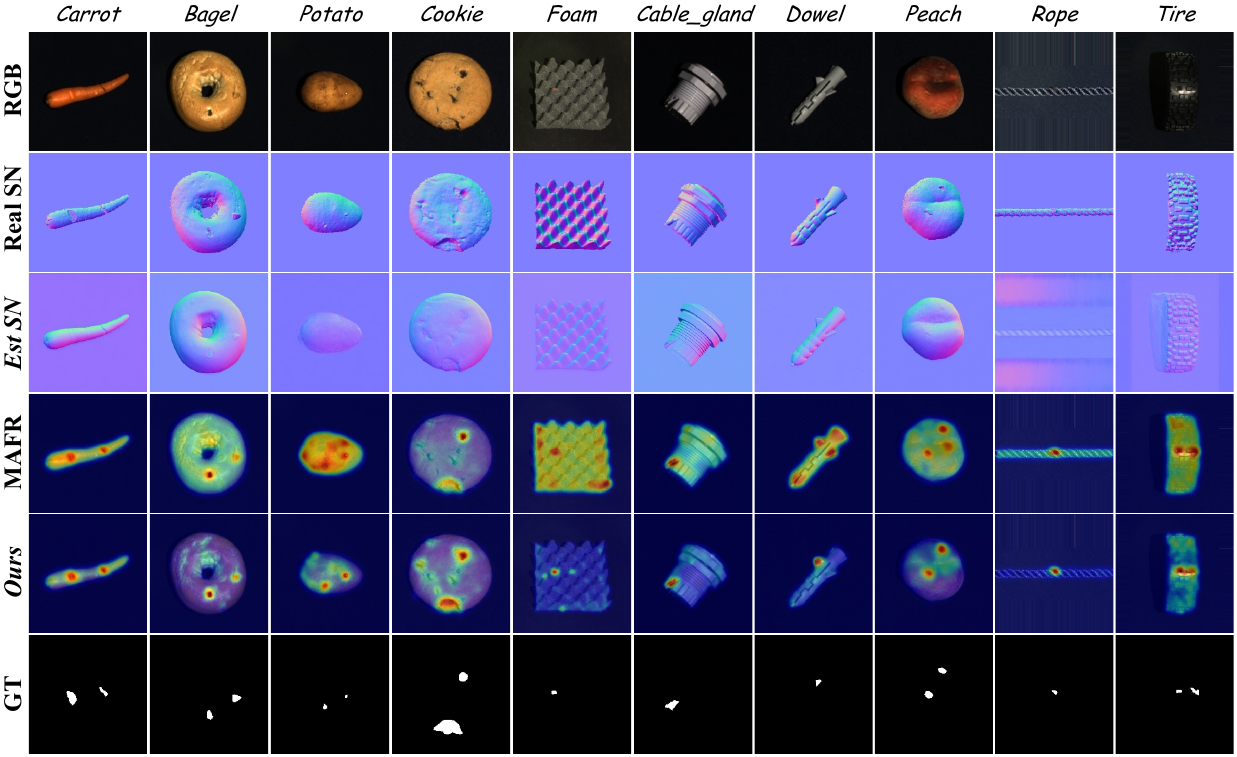}
    \caption{Extended qualitative results on the MVTec 3D-AD dataset (Part 1) under the 4-shot training setting. From top to bottom: RGB image, Real Normal map, Surface Normal map, Ground Truth mask, MAFR prediction, and our CMDS-AD predicted anomaly map.}
    \label{fig:mvtec_qual_1}
\end{figure}

\begin{figure}[htbp]
    \centering
    \includegraphics[width=0.91\columnwidth]{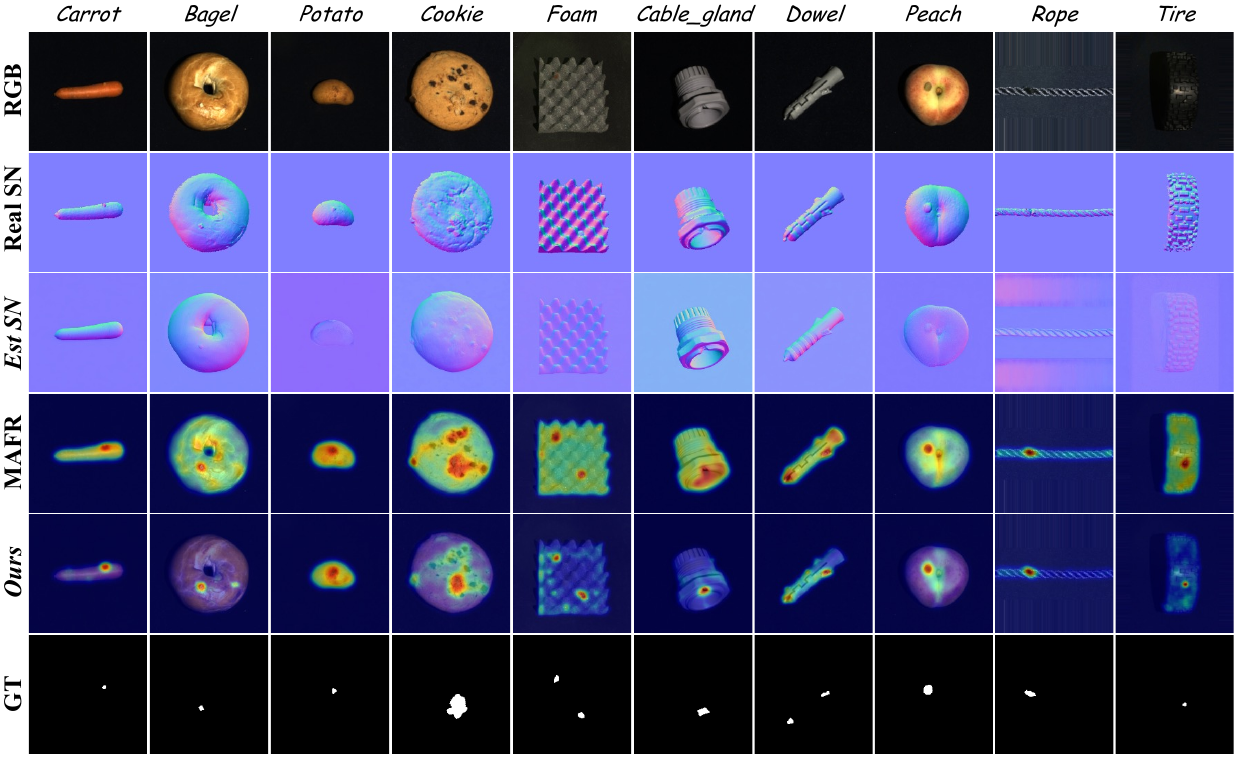}
    \caption{Extended qualitative results on the MVTec 3D-AD dataset (Part 2) under the 4-shot training setting. The layout of the visualizations is identical to Figure \ref{fig:mvtec_qual_1}.}
    \label{fig:mvtec_qual_2}
\end{figure}

\begin{figure}[b!]
    \centering
    \includegraphics[width=0.91\columnwidth]{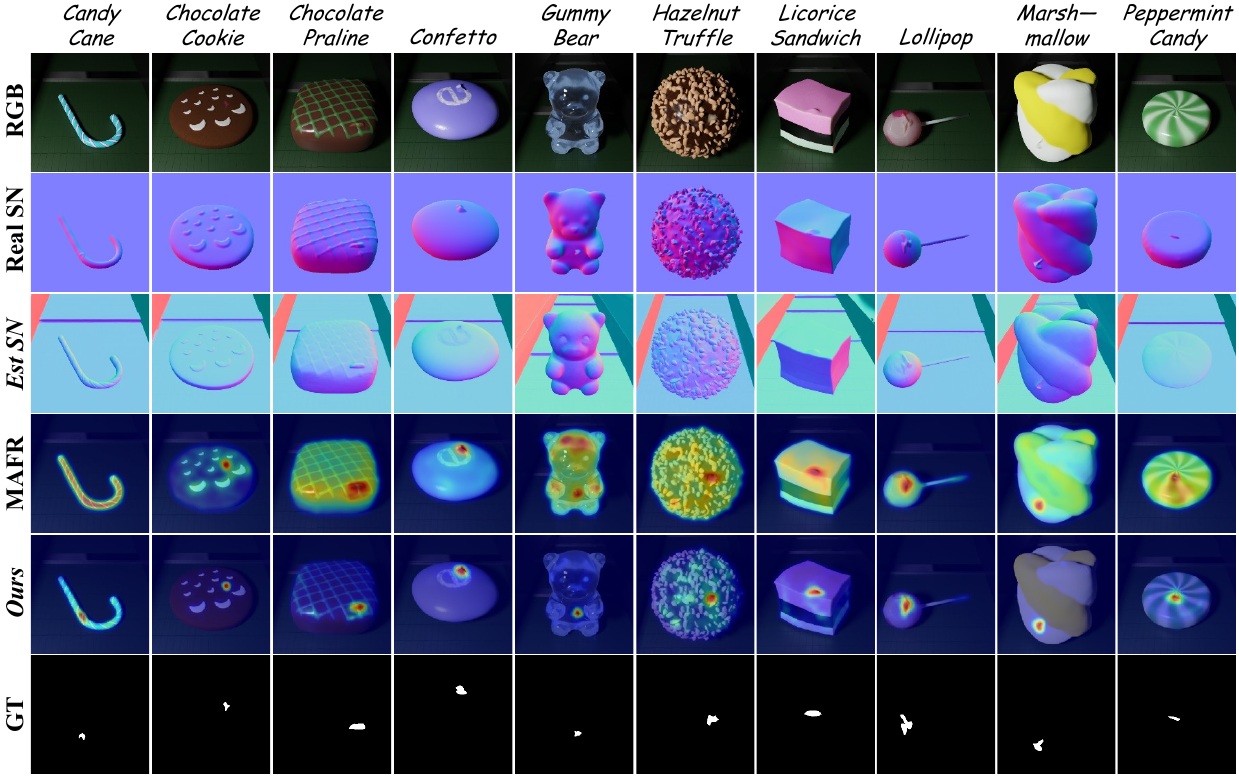}
    \caption{Extended qualitative results on the EyeCandies dataset (Part 1) under the 4-shot training setting. From top to bottom: RGB image, Real Normal map, Surface Normal map, Ground Truth mask, MAFR prediction, and our CMDS-AD predicted anomaly map.}
    \label{fig:eyecandies_qual_1}
\end{figure}

\begin{figure}[htbp]
    \centering
    \includegraphics[width=0.91\columnwidth]{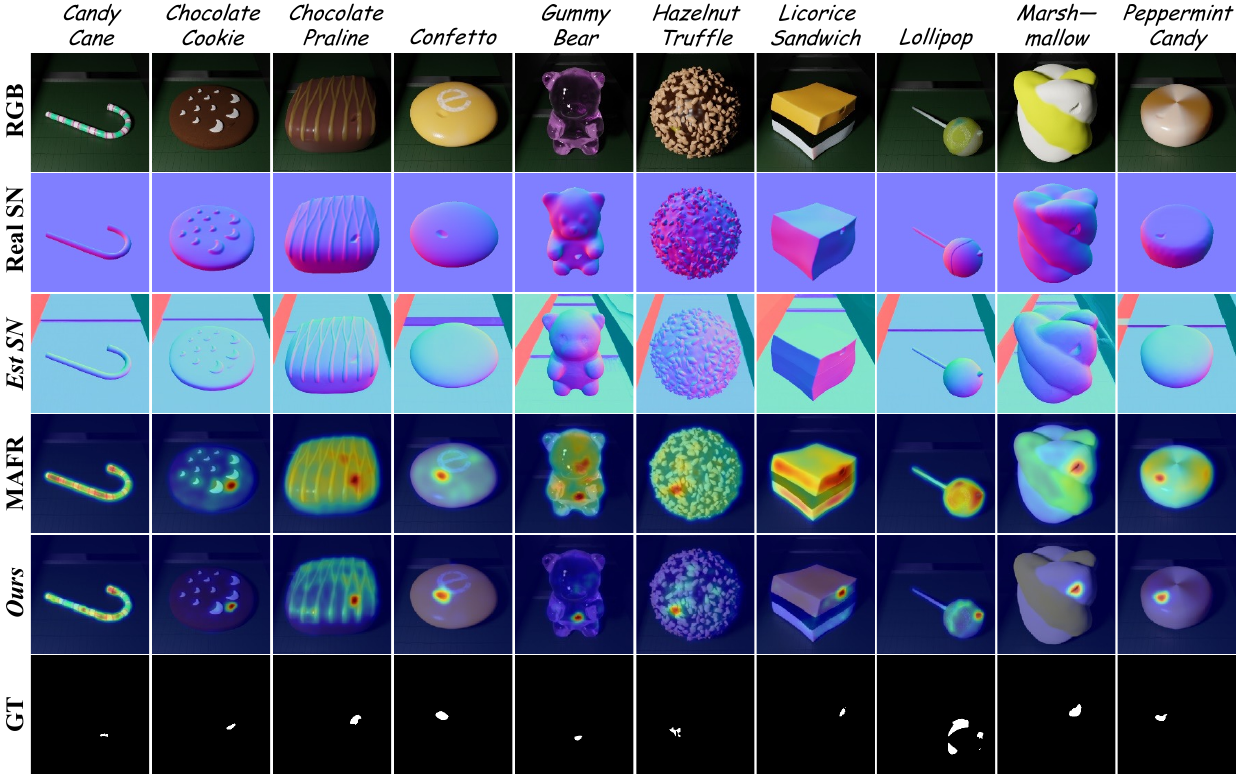}
    \caption{Extended qualitative results on the EyeCandies dataset (Part 2) under the 4-shot training setting. The layout of the visualizations is identical to Figure \ref{fig:eyecandies_qual_1}.}
    \label{fig:eyecandies_qual_2}
\end{figure}

To complement the visual demonstrations provided in the main text, we present an extended gallery of qualitative results covering the remaining categories of both the MVTec 3D-AD and EyeCandies datasets. Notably, all visual results presented in this section are generated under the 4-shot training setting. Specifically, Figures \ref{fig:mvtec_qual_1} and \ref{fig:mvtec_qual_2} display the anomaly localization results for the MVTec 3D-AD dataset, which features diverse industrial materials and complex 3D topologies. Meanwhile, Figures \ref{fig:eyecandies_qual_1} and \ref{fig:eyecandies_qual_2} demonstrate our performance on the EyeCandies dataset, which introduces challenging scenarios with intricate, highly reflective textures.

As illustrated in these figures, the visualizations are arranged from top to bottom, comprising the input RGB image, the Real Normal map, the Surface Normal map, the Ground Truth mask, the anomaly map predicted by MAFR, and our model's final dense prediction output. This comprehensive layout allows for a direct visual correlation between the raw multi-modal inputs, the distinct geometric cues provided by different normal representations, and the final localization performance. 

Compared to the MAFR baseline, which frequently struggles with over-segmentation and is easily distracted by benign high-frequency texture variations or sensor noise, our proposed CMDS-AD exhibits remarkable robustness. Benefiting from the dual-branch decoupling mechanism, our method effectively cross-verifies structural deviations and texture anomalies. Consequently, CMDS-AD not only accurately localizes subtle defects but also demonstrates exceptional precision in delineating complex anomaly boundaries across highly varied material surfaces, consistently suppressing spurious background activations and yielding predictions that closely align with the Ground Truth.

\subsection{Failure Cases Analysis}

\begin{figure}[htbp]
    \centering
    \includegraphics[width=1.0\columnwidth]{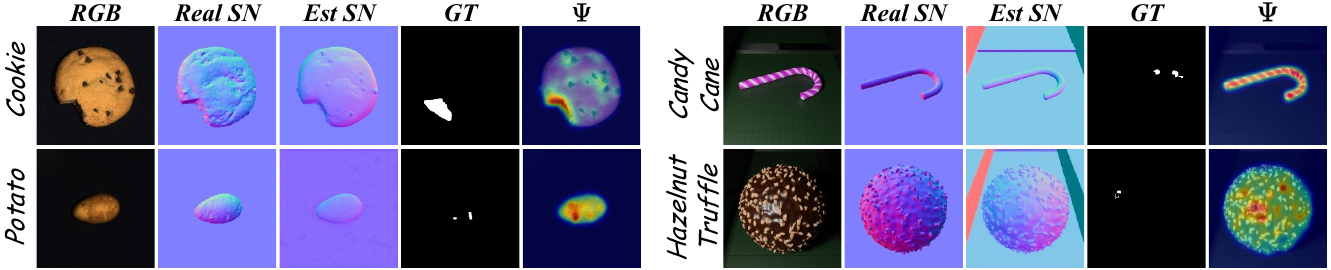}
    \caption{Typical failure cases of CMDS-AD. We highlight instances where the model misses specific structural defects or predicts broader false positive regions due to hyper-complex high-frequency patterns.}
    \label{fig:failure_cases}
\end{figure}

In Figure \ref{fig:failure_cases}, we highlight some typical failure cases of this approach under strictly constrained settings. For instance, in the first left row, we note that our method cannot accurately highlight the missing left part of the \textit{cookie}. This limitation fundamentally stems from the physical absence of the region itself: since the network cannot assign anomaly scores to non-existent pixels (or empty background), the anomaly signal is inevitably forced to manifest on the remaining broken edge adjacent to the defect. In the second left row, the \textit{potato} presents a tiny defect on its body, while the anomaly map—although covering the defect correctly—predicts a much broader anomaly region. This over-segmentation phenomenon can be attributed to the low-pass filtering nature of the estimation stream, where the highly localized defect signal bleeds into the smoothed surrounding features during the multi-scale feature comparison. In the first and second right rows, categories such as the \textit{candy cane} and the \textit{hazelnut truffle} present hyper-complex, high-frequency 2D or 3D patterns that produce higher anomaly scores compared to the real defects. Such false positives highlight a fundamental challenge in extreme few-shot scenarios: when the limited normal reference set fails to capture the full manifold of benign texture variations, the fusion mechanism may occasionally misinterpret these unseen but normal high-frequency patterns as structural anomalies.

% \clearpage  % TODO FINAL: This \clearpage needs to be removed from both review and camera-ready versions.

% \section*{Acknowledgements}
% Please insert your acknowledgments here.

% \bibliography{main}
 % 读入你的 Appendix.tex 文件

\end{document}